\documentclass[]{infiniai}
\usepackage{microtype}
\usepackage{amsfonts}
\usepackage[T1]{fontenc}   
\usepackage{lmodern}       
\usepackage{graphicx}
\usepackage{booktabs}
\usepackage{wrapfig}
\usepackage{listings}
\usepackage{amsmath}
\usepackage{amsthm}
\usepackage{amssymb} 
\usepackage{subcaption}
\usepackage{textcomp}

\usepackage{algpseudocode}
\usepackage[linesnumbered,lined,boxed,commentsnumbered,ruled,longend]{algorithm2e}

\usepackage{caption}
\usepackage{subcaption}
\definecolor{codebg}{RGB}{245,247,250}
\definecolor{codekw}{RGB}{0,92,175}   
\definecolor{codecm}{RGB}{120,120,120}
\definecolor{codestr}{RGB}{163,21,21} 
\definecolor{kwblue}{RGB}{40,40,180}
\definecolor{fnpurple}{RGB}{120,0,140}
\definecolor{commentgray}{RGB}{120,120,120}
\definecolor{algtitlebg}{HTML}{ECE3F7}
\newcommand{\algname}[1]{\colorbox{algtitlebg}{\textbf{#1}\strut}}

\lstdefinestyle{mystyle}{
    backgroundcolor=\color{codebg},
    basicstyle=\ttfamily\footnotesize,
    keywordstyle=\color{codekw}\bfseries,
    commentstyle=\color{codecm}\itshape,
    stringstyle=\color{codestr},
    frame=single,
    framesep=2pt,          
    rulecolor=\color{gray!40},
    xleftmargin=2pt,       
    xrightmargin=2pt,
    aboveskip=2pt,         
    belowskip=2pt,
    showstringspaces=false,
    breaklines=true,
    tabsize=2,
    emph={
        vFlow, block_size, Mean, GeMM, topK, Sum, attn, forward_cache, forward_indexer, Max, Min, Maximum, Conv, Softmax, Mask, Gather,
        HF, vLLM, vLLM_Omni, run_asr, run_llm, denoise, run_vae, run_tts, _run_liveavatar_s2v,
        run_conversational_agent
    },
    emphstyle=\color{fnpurple}\bfseries
}

\lstdefinestyle{custom}{
    backgroundcolor=\color{codebg},
    basicstyle=\ttfamily\footnotesize,
    frame=single,
    rulecolor=\color{black!20},
    breaklines=true,
    showstringspaces=false,
    tabsize=2,
    xleftmargin=4pt,
    xrightmargin=4pt,
    aboveskip=0pt,
    belowskip=0pt,
    columns=fullflexible,
    keepspaces=true,
    keywordstyle=\color{kwblue}\bfseries,
    commentstyle=\color{commentgray},
    morekeywords={function,for,if,in,continue,range},
    emph={
        program_id,num_programs,load,TILES,
        INIT_STATE,COMPUTE,COMPUTE_STEP,
        LOAD_BATCHED,LOAD_RAGGED,LOAD_PAGED,
        STORE_BATCHED,STORE_RAGGED,STORE_PAGED,
        layout,store_tensor,compute,load_tensor,load_tile,
        batched,ragged,paged,end_of_block
    },
    emphstyle=\color{fnpurple}\bfseries
}

\newcommand{\Sys}{\textsc{Vortex}\xspace}
\newcommand{\sys}{\textsc{Vortex}\xspace}
\newcommand{\vTensor}{\textsc{vTensor}\xspace}
\newcommand{\vtensor}{\textsc{vTensor}\xspace}
\newcommand{\vFlow}{\textsc{vFlow}\xspace}
\newcommand{\vflow}{\textsc{vFlow}\xspace}

\usepackage[capitalize,noabbrev]{cleveref}
\theoremstyle{plain}

\theoremstyle{definition}

\theoremstyle{remark}

\usepackage{enumitem}

\title{Vortex: Efficient and Programmable \\ Sparse Attention Serving for AI Agents}

\author{Zhuoming Chen$^1$, Xinrui Zhong$^2$, Qilong Feng$^1$, Ranajoy Sadhukhan$^1$, Yang Zhou$^1$, Michael Qizhe Shieh$^3$, Zhihao Jia$^1$, Beidi Chen$^1$}
\contact{zhuominc@andrew.cmu.edu, Xinrui.Zhong@rice.edu, \{qilongf, rsadhukh, yangzhuo6\}@andrew.cmu.edu, michaelshieh@nus.edu.sg, \{zhihaoj2, beidic\}@andrew.cmu.edu}
\affiliation{$^1$Carnegie Mellon University \\ $^2$Rice University \\ $^3$National University of Singapore}

\abstract{Sparse attention is becoming increasingly important for serving large language models (LLMs) as generation lengths continue to grow. However, deploying and evaluating \emph{new} sparse attention algorithms at scale remains highly engineering-intensive, slowing both \textbf{human researchers} and \textbf{AI agents} in exploring the sparse attention design. To address this challenge, we present \textsc{Vortex}, a system that combines a Python-embedded frontend language atop a page-centric tensor abstraction for expressing a broad range of sparse attention algorithms, with an efficient backend tightly integrated into modern LLM serving stacks. \textsc{Vortex} enables rapid prototyping, deployment, and evaluation of sparse attention algorithms, effectively translating their theoretical efficiency gains into real-world throughput improvements. As a result, \textsc{Vortex} substantially accelerates the design and iteration of sparse attention algorithms. \textbf{First}, AI agents use \textsc{Vortex} to automatically generate and refine diverse algorithms, the best reaching up to $3.46\times$ higher throughput than full attention while preserving accuracy. \textbf{Second}, \textsc{Vortex} extends sparse attention to emerging architectures and very large models that are otherwise hard to experiment with, reaching up to $4.7\times$ higher throughput on the MLA-based GLM-4.7-Flash and $1.37\times$ on the 229B-parameter MiniMax-M2.7 on NVIDIA B200 GPUs. 
}
\metadata[Github]{\url{https://github.com/Infini-AI-Lab/vortex_torch}}
\metadata[Website]{\url{https://infini-ai-lab.github.io/vortex_torch/}}
\metadata[Docs]{\url{https://infini-ai-lab.github.io/vortex_torch/docs}}

\begin{document}

\maketitle

\begin{figure}[h]
    \centering
     \subcaptionbox{Agentic Workflow with \sys\label{fig: workflow}}{
\includegraphics[width=0.36\linewidth]{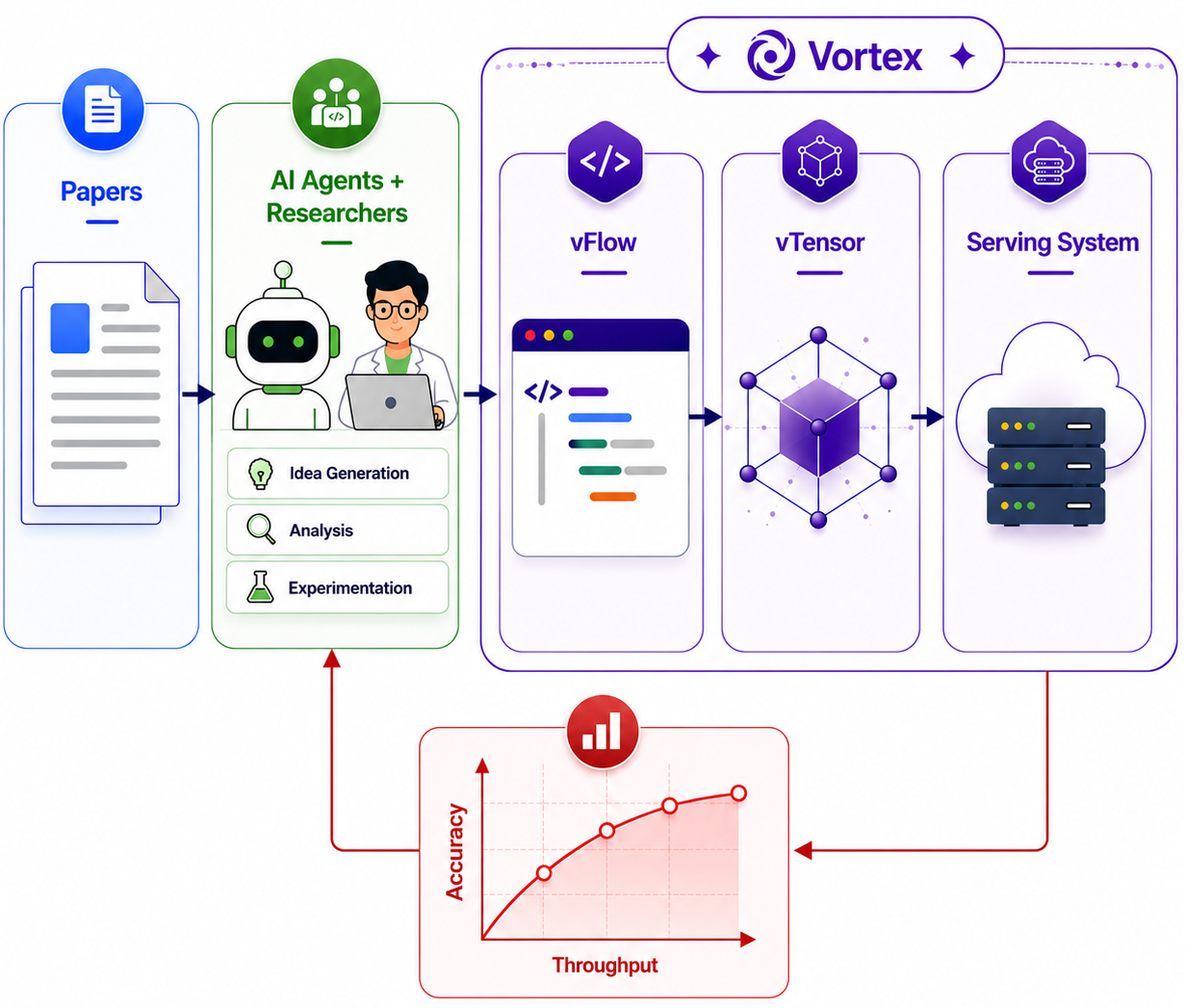}
    }
    \subcaptionbox{Agent-generated sparse attention\label{fig: allresutls}}{
\includegraphics[width=0.43\linewidth]{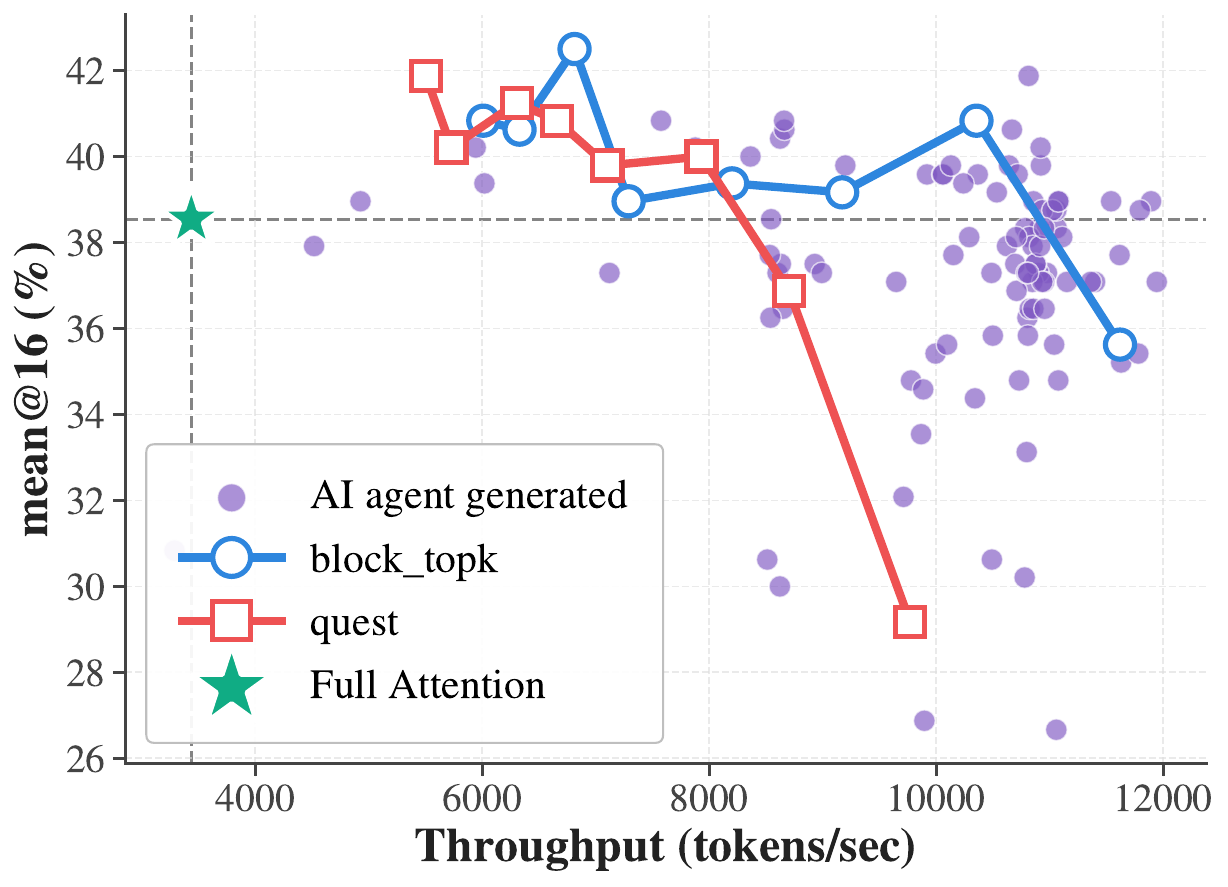}
    }
\caption{\textbf{(a). A workflow to study sparse attention algorithms using \sys.} \sys bridges sparse attention research and real-world serving systems by enabling easy programming and efficient inference for fast iteration and large-scale validation. \textbf{(b). Performance of agent-generated sparse attention algorithms on NVIDIA H200 SXM.} Each point represents one sparse attention generated or optimized by AI agents using \sys. By combining programmable abstractions with efficient execution in serving systems, \sys enables large-scale autonomous experimentation and iterative optimization. The best generated sparse attention achieves up to $3.46\times$ higher throughput than full attention while preserving accuracy.}
\end{figure}
\newpage

\section{Introduction}

Sparse attention is emerging as a fundamental technique for reducing inference costs in large language models (LLMs), driven by the rapid growth of generation lengths in applications such as reasoning~\citep{guo2025deepseek}, agentic systems~\citep{wang2024openhands}, and reinforcement learning~\citep{guo2025deepseek,schulman2017proximal}. In these workloads, key-value (KV) cache movement during decoding has become the primary system bottleneck. As a result, sparse attention is being adopted both as a core architectural component in state-of-the-art models such as DeepSeek~\citep{liu2025deepseek,yuan2025native,deepseekai2026deepseekv4} and GLM-5.1~\citep{zeng2025glm}, and as a drop-in optimization for pretrained models~\citep{chen2024magicpig,tang2024quest,Wang2026PrismSB,sun2024shadowkv,singhania2024loki,zhang2023h2o,yang2025lserve,lin2024infinite,xiao2024infllm,zhao2025infllm,cai2024lococo,yang2024tidaldecode}. To accelerate progress in sparse attention algorithms,  reducing implementation complexity while maintaining high serving efficiency is becoming critical.

\begin{wrapfigure}{r}{0.45\linewidth}
    \centering
    \includegraphics[width=\linewidth]{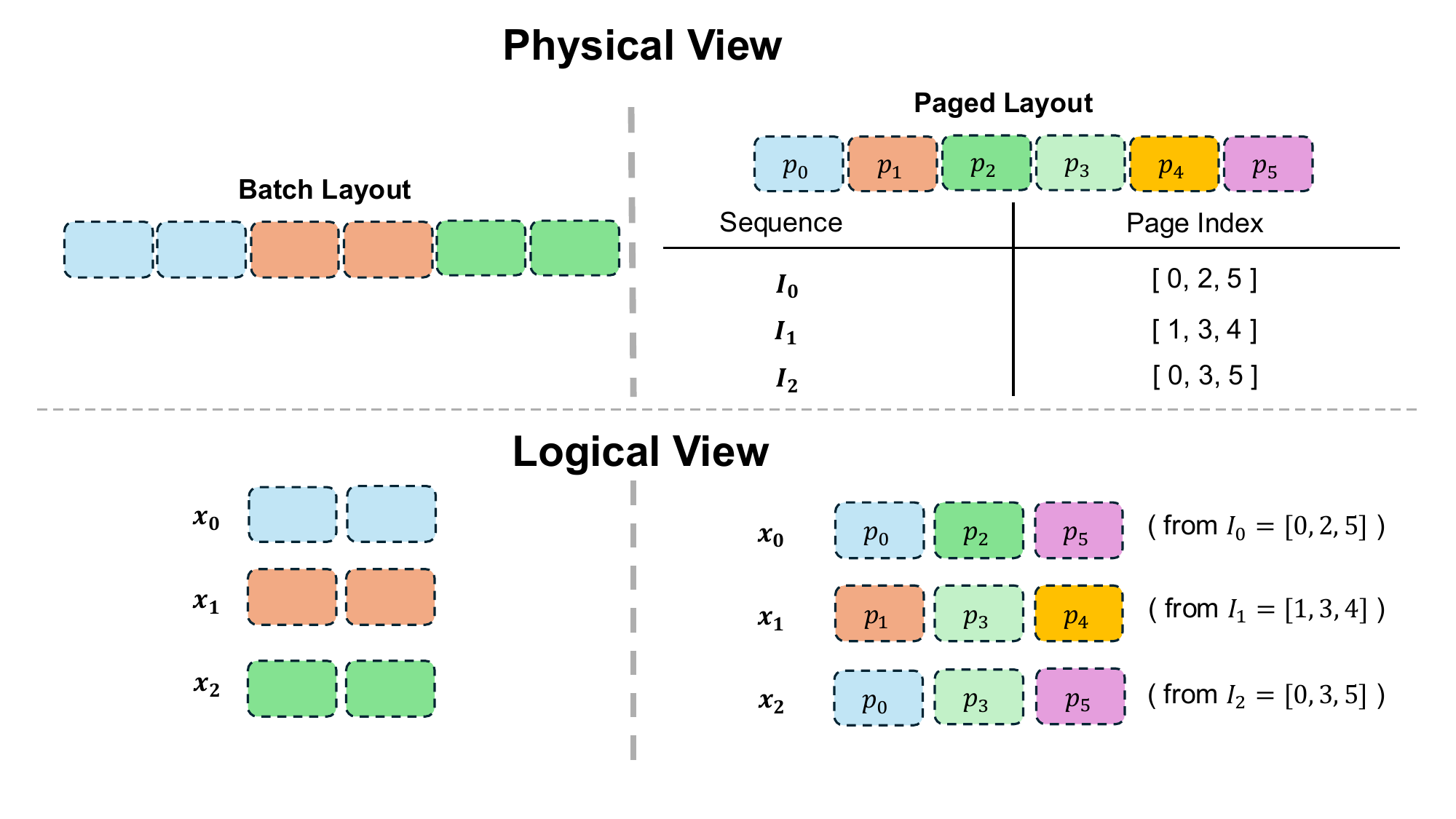}
    \caption{Illustration of paged layouts.}
    \label{fig:paged}
\end{wrapfigure}


Deploying and validating \emph{new} sparse attention algorithms at scale while achieving end-to-end speedups remains highly challenging and engineering-intensive, substantially slowing both \textbf{human researchers} and \textbf{emerging AI agents} in exploring the sparse attention design space. The core difficulty arises from modern LLM serving systems that adopt paged attention to reduce memory fragmentation, in which the KV cache is stored in a physically non-contiguous, block-sparse layout accessed through indirect addressing~\citep{vllm}. This breaks the memory assumptions underlying existing tensor frameworks such as PyTorch~\citep{paszke2019pytorch} for batch (contiguous) tensor layouts (illustrated in~\Cref{fig:paged}). Consequently, sparse attention algorithms are difficult to express within existing tensor abstractions. Despite substantial progress toward addressing these challenges, three key gaps remain.


\noindent \textbf{1. Difficult to Generalize to Dynamic Sparsity.} 
Prior work (e.g., FlashInfer~\citep{flashinfer}, FlexAttention~\citep{dong2024flex}) generalized the \emph{attention operator} to tensors with a paged layout. Thus, they can optimize sparse attention kernels once the sparsity patterns (i.e., the set of KV pairs each query attends to) are \emph{known}. While these approaches achieve substantial speedups for \emph{static sparse attention}, they do not naturally extend to \emph{dynamic sparse attention}, which proved to be more accurate. In dynamic settings, determining the sparsity pattern on the fly is a central component of the algorithm and incurs computational overhead that exceeds that of the attention operator.

\noindent \textbf{2. Lack of Programmability.}
Existing LLM serving systems~\citep{sglang,vllm,2023lmdeploy} support several sparse attention methods but provide limited programmability for integrating new algorithms. As a result, adding a new variant can require substantial engineering effort. For example, implementing Double Sparse~\citep{yang2024post} in SGLang~\citep{sglang} requires roughly 2000 lines of code, much of which reimplements operators such as \texttt{GeMM}, \texttt{Reduce}, and \texttt{Top-K} over paged tensors. This significantly hinders both researchers and AI agents~\citep{openevolve,novikov2025alphaevolve} from rapidly prototyping and iterating on new sparse attention ideas.



\noindent \textbf{3. Incompatibility with Evolving Serving Systems.} Many sparse attention methods rely on custom kernels~\citep{sun2024shadowkv,tang2024quest,lin2025twilight,chen2024magicpig}, but remain incompatible with key components of modern LLM serving stacks, including paged attention~\citep{vllm}, prefix caching~\citep{sglang}, and rapidly evolving attention backends~\citep{flashinfer,zadouri2026flashattention,shah2024flashattention,flashmla2025}. As a result, their theoretical speedups often fail to translate into practical end-to-end gains. For example, Quest's original implementation is $44.4\times$ slower than SGLang full attention, taking nearly two hours to evaluate Llama 3.1-8B on AMC23, despite optimized sparse kernels. These results suggest that evaluating sparse attention on long-generation workloads is impractical without tight integration with modern serving systems, even during prototyping and iteration.

An ideal framework should allow AI agents to express sparse attention mechanisms by specifying \emph{what} sparsity pattern to apply and \emph{how} attention is computed, while abstracting away from low-level tensor layouts and memory-management details. At the same time, the framework should transparently optimize execution and remain fully compatible with existing LLM serving infrastructures, enabling researchers to obtain rapid and realistic performance feedback.

A natural approach toward such a framework is to implement each sparse attention variant as a dedicated custom operator. However, this strategy fundamentally lacks compositionality: every new sparsity pattern or algorithm requires a separate implementation, making the system difficult to extend and unable to scale with the rapidly growing diversity of sparse attention designs.


Inspired by abstractions for sparse tensor linear algebra~\citep{abdelfattah2024interface,10.1145/567806.567810,10.1145/3582016.3582047,flashinfer}, which provide systematic representations and optimizations for sparse computation, we propose \vTensor, a \textbf{page-centric tensor system}. \vTensor generalizes sparse tensor computation to a paged layout, enabling sparse attention algorithms to be expressed in a flexible and composable manner without exposing low-level memory layouts or addressing details. However, leveraging \vtensor to program sparse attention and achieve speedups for LLM serving requires overcoming several technical challenges.
\begin{itemize}
[itemsep=0.0pt,topsep=0pt,leftmargin=*]
   \item \textbf{Simple and Modular Programming Interface.} The framework should provide an intuitive tensor view that hides low-level layout details and reduces user complexity. Sparse attention algorithms should also be expressed as modular components, even when they involve computation across stages of the serving pipeline.
   
   \item \textbf{Efficient Execution on Paged Layouts.} Supporting heterogeneous tensor layouts makes kernel implementation challenging. Operators must execute efficiently on non-contiguous memory layouts while minimizing gather operations and redundant data movement.
\end{itemize}

To address these challenges, we propose \sys, a system for efficient and programmable sparse attention serving. As shown in~\Cref{fig: workflow}, \sys consists of three components: (1) a \textbf{frontend language} (\vflow) embedded in Python for expressing sparse attention algorithms; (2) an \textbf{interpreter} that translates \vflow programs to executable \vtensor operators; and (3) an \textbf{execution backend} that integrates seamlessly with existing serving systems.

\begin{itemize}[itemsep=0.0pt,topsep=0pt,leftmargin=*]

\item In~\Cref{sec:running example}, we introduce the \vflow programming model through an example based on block top-$k$ attention, and demonstrate its expressiveness across a wide range of sparse attention algorithms in~\Cref{sec:generalizability}. 

\item In~\Cref{sec:vtensor}, we present the \vTensor abstraction for paged tensor computation. We then describe in~\Cref{sec:interandexec} how \vflow programs are compiled and executed efficiently as \vTensor programs within modern serving systems.

\item In~\Cref{sec:optimization}, we present three key optimizations in \sys: a workload planner for balanced GPU thread-block scheduling, kernel fusion to reduce intermediate memory traffic (see~\Cref{sec:planner}), and stochastic optimizations for radix top-$k$, a critical operator in sparse attention (see~\Cref{sec:fast topk}). In~\Cref{sec:backend}, we present \sys's compatibility with existing optimized attention backends, including GQA and MLA.

\end{itemize}

In~\Cref{sec:aiagents}, we first demonstrate that \sys enables rapid innovation and iteration of sparse attention algorithms with AI agents, as shown in~\Cref{fig: allresutls}. In~\Cref{sec:innovation}, Claude Code and Codex are prompted to generate a diverse set of sparse attention algorithms that achieve strong accuracy-throughput trade-offs. In~\Cref{sec:iteration}, an 18-hour autonomous optimization loop improves throughput by up to $3.46\times$ over full attention while preserving accuracy. In~\Cref{sec:efficiency}, we then evaluate the serving efficiency of \sys. Compared to SGLang, \sys achieves up to $3.60\times$ and $2.98\times$ throughput improvements over full attention for block top-$k$ and Quest, respectively, while maintaining accuracy. In terms of user latency, \sys reduces P95 latency by up to $11.7\times$ for block top-$k$ attention and $12.8\times$ for Quest under long-input workloads, all evaluated on NVIDIA \textsc{H200} SXM.

Beyond these, \sys readily extends to new attention architectures and to very large models that are otherwise hard to even experiment with. On the MLA-based GLM-4.7-Flash, running on an NVIDIA \textsc{B200} (\Cref{sec:mla}), a rope-aware sparse attention we design in \vflow achieves up to a $4.7\times$ speedup over full attention, and on the 229B-parameter MiniMax-M2.7 served across four NVIDIA \textsc{B200} GPUs (\Cref{sec:efficiency}), block top-$k$ achieves up to a $1.37\times$ speedup and even exceeds full-attention accuracy. These results show that \sys not only delivers high-performance sparse attention serving but also provides a scalable step towards automated research and optimization of sparse attention by AI agents.

\section{Related Work}
\label{sec:background}
\textbf{LLM Serving Systems.}
A large body of work has improved the efficiency of LLM serving~\citep{yu2022orca, vllm, patel2024splitwise, agrawal2024taming, zhong2024distserve, holmes2024deepspeed, tensorrtllm, qin2024mooncake, sglang, sheng2023flexgen, mei2024helix, oliaro2024flexllm, miao2023spotserve, miao2023towards,2023lmdeploy}. Orca~\citep{yu2022orca} introduced \emph{continuous batching}. vLLM addresses the memory fragmentation bottleneck with \emph{paged-attention}. SGLang~\citep{sglang} focuses on \emph{prefix caching}, which reuses KV cache across repeated prefixes to reduce redundant computation.

\noindent \textbf{Sparse Attention.}
Sparse attention for LLMs can be broadly divided into \emph{training-time} and \emph{inference-time} approaches. Training-time methods incorporate sparsity directly into model architectures during pre-training ~\citep{deepseekai2026deepseekv4,liu2025deepseek,yuan2025native,lu2025moba,huang2025nosa,NEURIPS2021_9185f3ec,zandieh2023kdeformer}.  Inference-time methods instead accelerate pretrained dense models with little or no retraining. These approaches are commonly categorized as \emph{static} or \emph{dynamic}. Static sparse attention~\citep{xiao2023efficient,agarwal2025gpt,xiaomi2025mimo,child2019generating,ding2023longnet} applies predefined sparsity patterns independent of input content, typically based on token positions. Dynamic sparse attention~\citep{tang2024quest,xiao2024infllm,chen2024magicpig,sun2024shadowkv,singhania2024loki,lin2025twilight,zhao2025infllm,gao2024seerattention,zhang2023h2o,MLSYS2024_bbb75065,ge2023model,kim2025kvzip,li2024snapkv,adnan2024keyformer} instead constructs input-dependent sparsity patterns by selecting KV entries according to the current query, hidden states, or content relevance.




\noindent \textbf{System Optimizations for Attention.}
Prior systems such as FlashAttention~\citep{flash_attn,flash_attn2,dao2022flashattention,zadouri2026flashattention}, Flash-Decoding~\citep{cascade-inference,hong2024flashdecodingfasterlargelanguage}, and SlimAttention~\citep{he2024inferenceperformanceoptimizationlarge} accelerate attention through improved hardware utilization and parallelism. Frameworks including FlashInfer~\citep{flashinfer}, FlexAttention~\citep{dong2024flex} further expose sparse attention interfaces based on KV block indices or masks for paged attention. However, these systems generally assume the sparse KV indices are already available, leaving the challenge of efficiently constructing them unresolved. More recent systems, such as LServe~\citep{yang2025lserve} and SparseServe~\citep{zhou2025sparseserve}, explore end-to-end sparse attention optimization but focus on specific algorithms or deployment settings, including Quest~\citep{tang2024quest} and block top-$k$ attention under memory-constrained workloads. 

\emph{Currently, \sys focuses on the decoding phase, where KV memory is the primary bottleneck, and leaves the prefilling phase for future work.}

\section{Programming Model: \vflow}
\label{sec:programming model}


In this section, we present the programming model of \Sys, namely \vflow, a unified abstraction for expressing sparse attention algorithms. We begin in~\Cref{sec:running example} with a concrete case study on block top-$k$ attention to illustrate the core design principles of \vflow. In particular, \vflow adopts a single-request mental model, treating tensors as batch size 1, while exposing flexible and composable primitives for constructing sparse attention mechanisms. We then show in~\Cref{sec:generalizability} how this abstraction naturally generalizes to a broad range of sparse attention algorithms.


\subsection{A Running Example: Block Top-$k$ Attention}
\label{sec:running example}

Let $\boldsymbol{q} \in \mathbb{R}^{g \times d}$ denote the query, and let 
$\boldsymbol{K} \in \mathbb{R}^{n \times d}$ and 
$\boldsymbol{V} \in \mathbb{R}^{n \times d}$ 
denote the key and value caches, respectively. We partition the sequence into $B = \lfloor n / b \rfloor$ non-overlapping blocks of size $b$. The mathematical formulation for block top-$k$ attention (for a single key value head) is defined in~\Cref{fig:mathofblocktopk} and the corresponding {\vFlow} implementation is presented in~\Cref{fig:vflowofblocktopk}, where primitives by {\vFlow} are shown in {\color{fnpurple}purple}.

This formulation naturally decomposes into two stages with distinct computational characteristics:

\textbf{(1) Query-independent preprocessing.}
\Cref{eq:block_topk_1} computes block-level key summaries $\widetilde{\boldsymbol{k}}_j$ by averaging keys within each block. 
These summaries depend only on the key cache $\boldsymbol{K}$ and are independent of the query, and thus can be precomputed once during cache construction and reused across decoding steps. This stage corresponds to \texttt{forward\_cache()} in~\Cref{fig:vflowofblocktopk}, where \vflow presents a \emph{logical view} of \texttt{c["centroids"]} as a contiguous tensor of shape \texttt{[1, B, d]}. 

\textbf{(2) Query-dependent execution.}
\Cref{eq:block_topk_2,eq:block_topk_4,eq:block_topk_5} define the dynamic computation performed for each incoming query $\boldsymbol{q}$. 
Specifically, the model (i) computes block relevance scores $\widetilde{s}_j$, (ii) selects the top-$k$ blocks, and (iii) performs attention over the corresponding tokens. This stage must be executed per token, as both the selected index set $\mathcal{I}$ and the output $\boldsymbol{o}$ depend on the current query.  This stage is implemented as \texttt{forward\_indexer()}, where \vflow exposes \emph{logical contiguous views} for all tensors: \texttt{c["centroids"]} as \texttt{[1, B, d]}, \texttt{c["k"]} and \texttt{c["v"]} as \texttt{[1, B, b, d]}, and \texttt{q} as \texttt{[1, g, d]}. Under this logical view, users can easily reason about intermediate tensor shapes; for example, \texttt{s} $\in \mathbb{R}^{1 \times B \times g}$ and \texttt{i} $\in \mathbb{R}^{1 \times 1 \times k}$. 

\begin{figure}[t]
\centering
\begin{minipage}[b]{0.47\linewidth}
{\fontsize{8pt}{9pt}\selectfont
\setlength{\jot}{2pt}
\begin{align}
\widetilde{\boldsymbol{k}}_j &= \operatorname{mean}\!\left(\boldsymbol{K}_{[jb:(j+1)b-1]}, \text{dim=0}\right), j \in [0, B) \label{eq:block_topk_1} \\
\widetilde{s}_j &= \operatorname{mean}\!\left(\boldsymbol{q} \widetilde{\boldsymbol{k}}_j^T, \text{dim=0}\right), j \in [0, B) \label{eq:block_topk_2} \\
\mathcal{I} &= \left\{ i \in [0,n) \mid \left\lfloor \frac{i}{b} \right\rfloor
\in \operatorname*{arg\,top\text{-}k}_{j \in [0,B)} \widetilde{s}_j \right\} \label{eq:block_topk_4} \\
\boldsymbol{o} &= \operatorname{softmax}\!\left(\frac{\boldsymbol{q}\boldsymbol{K}_{\mathcal{I}}^T}{\sqrt{d}}\right)\boldsymbol{V}_{\mathcal{I}} \label{eq:block_topk_5}
\end{align}
}
\captionsetup{justification=centering,singlelinecheck=false}%
\captionof{figure}{Formulations of block top-$k$ attention.}
\label{fig:mathofblocktopk}
\end{minipage}
\hfill
\begin{minipage}[b]{0.51\linewidth}
\begin{lstlisting}[language=Python]
class BlockTopK(vFlow):
    def forward_cache(c):
        c["centroids"] = Mean(c["k"], dim=1)

    def forward_indexer(q, c):
        s = GeMM(c["centroids"], q.T)
        i = topK(Mean(s, dim=2), dim=1)
        return attn(q, c["k"], c["v"], i)
\end{lstlisting}
\captionsetup{justification=centering,singlelinecheck=false}%
\captionof{figure}{Block top-$k$ attention in {\vFlow}.}
\label{fig:vflowofblocktopk}
\end{minipage}
\end{figure}

\subsection{Generalizability}
\label{sec:generalizability}
The \vflow model is highly general and can express a wide range of sparse attention algorithms. At a high level, its two-stage decomposition mirrors the standard paradigm in retrieval and similarity search~\citep{douze2025faiss,jegou2010product,malkov2018efficient,shrivastava2014asymmetric}: query-independent auxiliary structures (e.g., summaries, centroids) are constructed offline to enable efficient query-time selection. We further demonstrate the expressiveness of \vflow by answering the following questions. 

\textbf{Q1. Can \vFlow easily express more complicated dynamic sparse attention algorithms?} 
Yes. \vflow can naturally express a wide range of dynamic sparse attention algorithms by composing a small set of reusable operators. As shown in the examples (\texttt{DoubleSparse}~\citep{yang2024post}, \texttt{BlockTopK\_NSA}~\citep{yuan2025native}, \texttt{Quest}~\citep{tang2024quest}, and \texttt{H2O}~\citep{zhang2023h2o}, in~\Cref{fig:ds,fig:nsa,fig:quest,fig:h2o}.), different methods can be implemented by combining common building blocks such as \texttt{GeMM}, \texttt{Softmax}, \texttt{Top-K}, \texttt{Gather}, and reduction operators (the full operator set is listed in~\Cref{tab:ops}). This compositionality is enabled by \vtensor (see~\Cref {sec:vtensor}), which provides a unified interface for logical tensors while handling physical layout transparently. 

\textbf{Q2. Can intermediate computation results of \texttt{forward\_indexer()} be cached?} 
Yes. \vflow enables caching of intermediate results by allowing operator outputs to be stored in a named cache, provided the user explicitly declares the cache in \texttt{forward\_cache()}. A concrete example is \texttt{H2O}, where \texttt{c["acc"]} accumulates attention scores across decoding steps. This design facilitates the implementation of algorithms that require runtime statistics~\citep{zhang2023h2o,yang2024tidaldecode,bai2026indexcache}.

\textbf{Q3. Can \vflow extend to static sparse attention?} 

Yes. Static sparse attention can be naturally expressed in \vflow by constructing index sets that depend only on sequence positions rather than input content. For example, one can obtain the current sequence length via \texttt{B = c["k"].shape[1]} and derive the token position accordingly. The sparsity pattern can then be explicitly specified by constructing the index set $i$ (e.g., \texttt{i = List[...]}), independent of the query values.

\begin{minipage}[b]{0.48\linewidth}
\begin{lstlisting}[language=Python]
class DoubleSparse(vFlow):
    def forward_cache(c):
        c["channel"]=Gather(c["k"])
    def forward_indexer(q,c):
        s=GeMM(c["channel"],
        Gather(q).T)
        i = topK(Sum(s))
\end{lstlisting}
\captionsetup{justification=centering,singlelinecheck=false}%
\captionof{figure}{Double Sparse}
\label{fig:ds}
\end{minipage}
\hfill
\begin{minipage}[b]{0.48\linewidth}
\begin{lstlisting}[language=Python]
class BlockTopK_NSA(vFlow):
    def forward_cache(c):
        c["centroids"]=Mean(c["k"])
    def forward_indexer(q,c):
        s=Softmax(c["centroids"]@q.T)
        i=topK(Sum(Conv(s))
\end{lstlisting}
\captionsetup{justification=centering,singlelinecheck=false}%
\captionof{figure}{NSA (block top-$k$ part)}
\label{fig:nsa}
\end{minipage}

\begin{minipage}[b]{0.48\linewidth}
\begin{lstlisting}[language=Python]
class Quest(vFlow):
    def forward_cache(c):
        c["max"]=Max(c["k"])
        c["min"]=Min(c["k"])
    def forward_indexer(q,c):
        s=Maximum(q*c["max"],
                  q*c["min"])
        i=topK(Max(Sum(s)))
\end{lstlisting}
\captionsetup{justification=centering,singlelinecheck=false}%
\captionof{figure}{Quest}
\label{fig:quest}
\end{minipage}
\hfill
\begin{minipage}[b]{0.48\linewidth}
\begin{lstlisting}[language=Python]
class H2O(vFlow):
    def forward_cache(c):
        c["acc"]=0
    def forward_indexer(q,c):
        s=Softmax(GeMM(c["k"],q.T))
        c["acc"]+=s
        i = topK(Sum(s))
        Mask(~i, c["acc"], -inf)
\end{lstlisting}
\captionsetup{justification=centering,singlelinecheck=false}%
\captionof{figure}{H$_2$O}
\label{fig:h2o}
\end{minipage}

\section{Interpretation: \vtensor}
\label{sec:interpretation}

The \vflow programs provide a simple yet effective logical tensor abstraction that allows programmers to assume tensors are contiguous and processed with an effective batch size of 1. However, tensors in real LLM serving systems are often stored using more complex layouts to support continuous batching and paged KV caches (\Cref{sec:detailed memory layouts}). Thus, we need an underlying tensor system that explicitly accounts for paged layouts to interpret the semantics of \vflow into executable programs. To ensure the flexibility and simplicity of \vflow, the tensor system must satisfy two key properties:

\textit{Compositional.} Complex algorithms should be constructed by composing a small set of primitive tensor operators, rather than relying on highly specialized monolithic kernels. This design improves modularity, extensibility, and portability across different serving workloads.

\textit{Self-Contained.} The outputs of tensor operators should remain in formats directly consumable by subsequent operators. In particular, intermediate results should avoid unnecessary materialization or layout conversion, enabling efficient operator chaining on paged tensors.

Therefore, we introduce a new abstraction, \vtensor, which extends PyTorch tensors with explicit layout information. In~\Cref{sec:vtensor}, we present the definition of \vtensor and its operator. In~\Cref{sec:interandexec}, we present how to interpret \vflow programs into lower-level \vtensor functions, which are executable in serving systems.

\subsection{Definitions}
\label{sec:vtensor}

\textbf{Tensor Definitions.}
Inspired by~\citet{flashinfer}, we first introduce a unified abstraction for representing tensors with different memory layouts. A \vtensor is defined as
\begin{equation}
    \boldsymbol{x}^{v} = (\boldsymbol{x}, \mathcal{C})
    \label{eq:tensordef}
\end{equation}
where $\boldsymbol{x}$ denotes the underlying Pytorch tensor and $\mathcal{C}$ captures layout metadata defined as $\mathcal{C} = (b, \mathbf{p}, \mathbf{I})$, where $b$ is the batch size, $\mathbf{p}$ is an index pointer array, and $\mathbf{I}$ is an index structure.  In the paged layout, both $\mathbf{p}$ and $\mathbf{I}$ are present, where $\mathbf{p}$ defines sequence-level offsets and $\mathbf{I}$ specifies the mapping from sequences to page indices in a ragged form.

\textbf{Operator Definitions.}
The operators of \vtensor are defined by applying standard PyTorch operators independently to each sequence within the batch. Consider input tensors $\boldsymbol{x}^{v}_0, \boldsymbol{x}^{v}_1, \dots, \boldsymbol{x}^{v}_{n-1}$ that share the same batch size $b$, and a PyTorch operator $\mathcal{F}$ producing $k$ outputs. For each sequence $0 \le i < b$, the operator is evaluated as
\(
\boldsymbol{y}^{v}_{0,i}, \boldsymbol{y}^{v}_{1,i}, \dots, \boldsymbol{y}^{v}_{k-1,i}
=
\mathcal{F}(
\boldsymbol{x}^{v}_{0,i},
\boldsymbol{x}^{v}_{1,i},
\dots,
\boldsymbol{x}^{v}_{n-1,i}
).
\)

The resulting tensors are then aggregated into output \vtensor objects $\boldsymbol{y}^{v}_0, \dots, \boldsymbol{y}^{v}_{k-1}$ through shape and layout propagation. We present the details in~\Cref{sec:discussion}.

 Based on this abstraction, we implement a broad set of tensor operators and build a compiler that automatically performs kernel synthesis and operator fusion (see~\Cref{sec:planner}).

\subsection{Lowering \vflow to \vtensor}
\label{sec:interandexec}
In this section, we present how \vflow programs are translated into sequences of \vtensor operators by interpreting operator semantics and specifying tensor layout information (i.e., $\mathcal{C}$ in \Cref{eq:tensordef}).

\textbf{Operator Interpretation.}
Each variable in \vflow is interpreted as a \vtensor. For example, the statement
\(
\texttt{s = GeMM(c["centroids"], q.T)}
\)
is interpreted as
\(
\boldsymbol{s}^{v}_i
=
\texttt{GeMM}\big(
\texttt{c["centroids"]}^{v}_i,
(\boldsymbol{q}^{v}_i)^T
\big),
\ 0 \le i < b,
\)
where $\boldsymbol{q}^{v}_i \in \mathbb{R}^{g \times d}$ and $\texttt{c["centroids"]}^{v}_i \in \mathbb{R}^{B_i \times d}$. The outputs $\boldsymbol{s}^{v}_i \in \mathbb{R}^{B_i \times g}$ are aggregated into a ragged \vtensor with pointer array $\mathbf{p}$ satisfying \(\mathbf{p}[i+1] = \mathbf{p}[i] + B_i\), yielding a tensor of shape \(\left(\sum_i B_i\right)\times g\).

\textbf{Tensor Layouts.}
Tensor layouts in \sys are either inherited from the serving system or assigned by the compiler. \texttt{q} uses standard batch layouts, while cache tensors such as \texttt{c["k"]} and \texttt{c["v"]} use paged layouts managed by the serving runtime. Named caches in \sys, e.g., \texttt{c["acc"]} and \texttt{c["centroids"]}, are also stored as paged \vtensor objects, whereas temporary intermediates use ragged layouts by default. By sharing the same paging and allocation mechanism as the KV cache, \sys naturally preserves compatibility with optimizations such as prefix caching.

\section{Execution Optimizations}
\label{sec:optimization}



While \vTensor provides a unified abstraction over heterogeneous tensor layouts, efficient execution still requires additional system-level optimizations. To address this, \sys incorporates several compiler and runtime optimizations. First, \sys plans workloads according to tensor layouts and data dependencies to improve execution efficiency. Second, it applies kernel fusion to reduce intermediate memory traffic and improve locality (see~\Cref{sec:planner}). Third, it introduces stochastic optimizations for radix top-$k$ to accelerate sparse attention indexing (see~\Cref{sec:fast topk}). Finally, to execute the selected sparse blocks efficiently, \sys builds on existing, highly optimized attention backends across attention variants and precisions, complementing them with a custom MLA decode kernel for geometries they do not cover, so that it automatically supports diverse models (see~\Cref{sec:backend}).

\subsection{Workload Planning and Kernel Fusion}
\label{sec:planner}

At the beginning of each decoding iteration, the batch size and sequence lengths are known, allowing \sys to plan workloads and generate optimized GPU kernels~\citep{flashinfer,cheng2025mirage}. Different operators follow different execution templates depending on their computation patterns. For example, operators such as \texttt{Softmax} and \texttt{Top-K} require specialized kernels, while operators including \texttt{GeMM}, elementwise operations, and reductions share a unified chunk-based execution template as shown in~\Cref{lst:seq-local-template}.

Based on the templates, \sys further performs kernel fusion to reduce intermediate memory traffic. Computation is represented as a directed acyclic graph, and operators sharing compatible execution templates are greedily fused to eliminate unnecessary intermediate reads and writes.
\subsection{Optimization for Radix Top-$k$}
\label{sec:fast topk}

As a non-matmul operator, top-$k$ selection becomes a bottleneck in \texttt{forward\_indexer}. A radix-based top-$k$ algorithm that iteratively partitions values into bins based on their radix digits is widely adopted. Its performance, however, is highly sensitive to the score distribution. In particular, when many large values fall into the same radix bin, the algorithm must repeatedly refine a large candidate set, leading to highly variable runtime.

To mitigate this issue, we introduce \emph{stochastic early termination}, which is a lossy variant that trades exactness for speed. Instead of fully refining the threshold bin, the algorithm terminates early once enough high-confidence candidates have been collected and then randomly samples the remaining elements from the threshold bin. This stochastic relaxation significantly reduces refinement overhead while providing a tunable trade-off between accuracy and performance.

\subsection{Attention Backend Compatibility: GQA and MLA}
\label{sec:backend}

To turn sparse selection into end-to-end speedups, \sys reuses existing, highly optimized attention backends rather than re-implementing attention, feeding the sparse block table from \texttt{forward\_indexer} into a paged decode kernel. For grouped-query attention (GQA), \sys supports the \texttt{flashinfer}~\citep{flashinfer} and \texttt{trtllm\_mha} (TensorRT-LLM) backends; for multi-head latent attention (MLA), used by DeepSeek~\citep{liu2025deepseek} and GLM~\citep{zeng2025glm}, it supports \texttt{trtllm\_mla}. Reusing these backends also inherits their precision support, such as fp8 KV cache and weights.

Vendor MLA kernels, however, are tied to a fixed geometry: \texttt{trtllm\_mla} assumes a specific latent shape and supports only block sizes of 32 or 64. \sys therefore adds \texttt{cuda\_mla}, its own MLA decode kernel supporting general latent geometries and finer blocks (down to 16), so the block granularity can match the sparsity budget. Together, these backends let \sys automatically support diverse models.

\section{Evaluation}
\label{sec:evaluation}
We evaluate \sys to answer the following two questions:
\begin{itemize}[itemsep=0.0pt,topsep=0pt,leftmargin=*]

\item \textbf{Q1. Is \sys really helpful in innovating and iterating sparse attention algorithms?} Yes, and we show it along three axes.
\begin{itemize}[itemsep=1.0pt,topsep=1.0pt,leftmargin=1.2em]
\item \emph{Innovation and iteration by AI agents} (\Cref{sec:innovation,sec:iteration}). Claude Code and Codex generate a broad set of structurally distinct sparse attention algorithms with Pareto-efficient accuracy-throughput trade-offs, and an 18-hour autonomous loop improves AIME24 throughput by up to $3.46\times$ over dense attention for Qwen3-1.7B while preserving accuracy.
\item \emph{Design for new architectures} (\Cref{sec:mla}). Multi-head latent attention (MLA) is a new architecture adopted by frontier models such as DeepSeek and GLM; \sys lets us rapidly design a rope-aware sparse attention for it, a $4\times$ end-to-end speedup over full attention at matched accuracy on GLM-4.7-Flash.
\item \emph{New understanding} (\Cref{sec:understanding}). \sys acts as a research instrument that reveals where the routing signal lives, identifying which query-key channels and which MLA components are critical.
\end{itemize}

\item \textbf{Q2. Is \sys making sparse attention algorithms really faster in deployment?} In~\Cref{sec:end-to-end}, for \textbf{server throughput}, \sys achieves up to $3.60\times$ and $2.98\times$ higher throughput than full attention for block top-$k$ and Quest, respectively, while maintaining accuracy. This advantage holds at the largest scale we test; on the 229B-parameter MiniMax-M2.7 served across four NVIDIA B200 GPUs with tensor parallelism (TP$=4$), block top-$k$ reaches up to $1.37\times$ higher throughput than full attention with the same accuracy. In~\Cref{sec:latency}, for \textbf{user latency}, \sys reduces P95 latency by $11.7\times$ for block top-$k$ and $12.8\times$ for Quest at a 16K prompt length under a request rate of 8.0.
\end{itemize}
Besides, we present the ablation study of the kernel efficiency in~\Cref{sec:ablations}.

\subsection{Researching Sparse Attention with AI Agents}
\label{sec:aiagents}
In this section, we present two use cases of \sys for AI-assisted research on sparse attention.



\textbf{Setup.} Our target model is Qwen3-1.7B deployed on NVIDIA H200 SXM GPU. We evaluate both reasoning accuracy and serving efficiency using pass@16 on AMC23 and AIME24 (\# Max Gen Tokens = 16K), as well as decoding throughput. To support AI-agent-driven research, we provide the agents with 8 representative sparse attention papers as references, along with the \sys documentation. The template for AI agents to submit algorithms is present in~\Cref{sec:templates}.

\subsubsection{Experiment 1: Algorithm Innovation} 
\label{sec:innovation}
We task Claude Code~\citep{anthropic2025claudecode} (Opus-4.7 and Sonnet-4.6) and Codex~\citep{openai2025codex} (GPT-5.5~\citep{openai2025gpt5}) with proposing and implementing 20 sparse attention algorithms each in a one-shot setting, explicitly prompting them to generate novel designs; we describe all of the proposed algorithms in~\Cref{sec:agent-algos}. This experiment evaluates whether AI agents can directly produce diverse and effective sparse attention algorithms when equipped with \sys.

\begin{figure}[t]
    \centering
    \includegraphics[width=\linewidth]{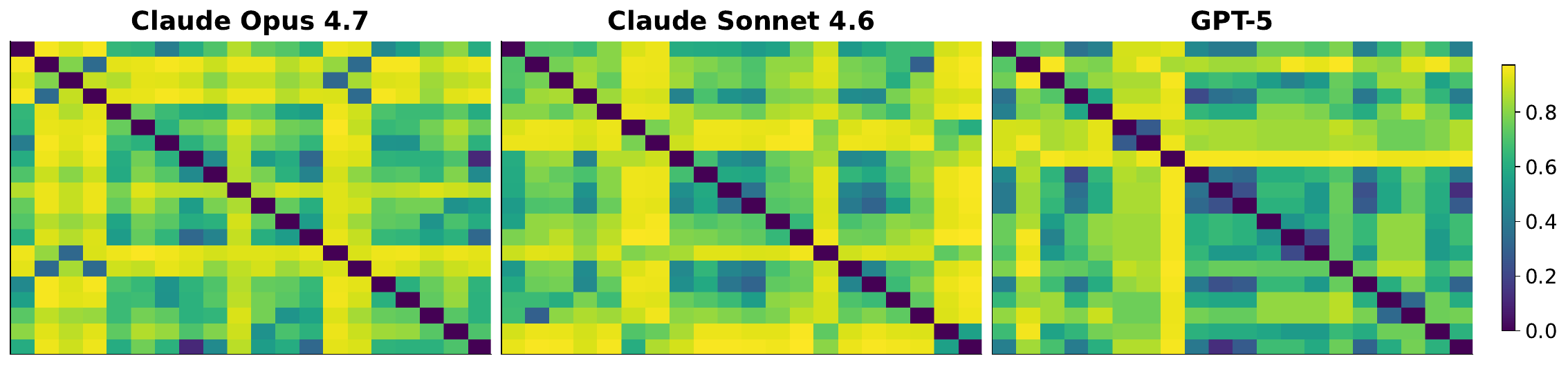}
    \caption{\textbf{Structural diversity of AI-agent-generated sparse attention algorithms.} Higher values indicate more distinct implementations. Claude Sonnet 4.6 achieves the highest overall diversity.}
    \label{fig:diversity}
\end{figure}

\textbf{Algorithm Diversity.}
As shown in~\Cref{fig:diversity}, we measure diversity using a structural distance metric computed from the generated implementations. For each program, we extract (1) cache definitions, (2) instantiated operators, and (3) operator invocation patterns using abstract syntax tree parsing, and compute pairwise Jaccard-style distances over these components. Claude Sonnet 4.6 achieves the highest average pairwise distance ($0.789$), followed by Claude Opus 4.7 ($0.770$) and GPT-5 ($0.709$). All models produce highly diverse implementations, indicating that AI agents can generate substantially different sparse attention designs in a single attempt when paired with \sys.

\begin{figure*}
    \centering
    \subcaptionbox{RULER\label{fig: rulerinno}}{%
\includegraphics[width=0.29\linewidth]{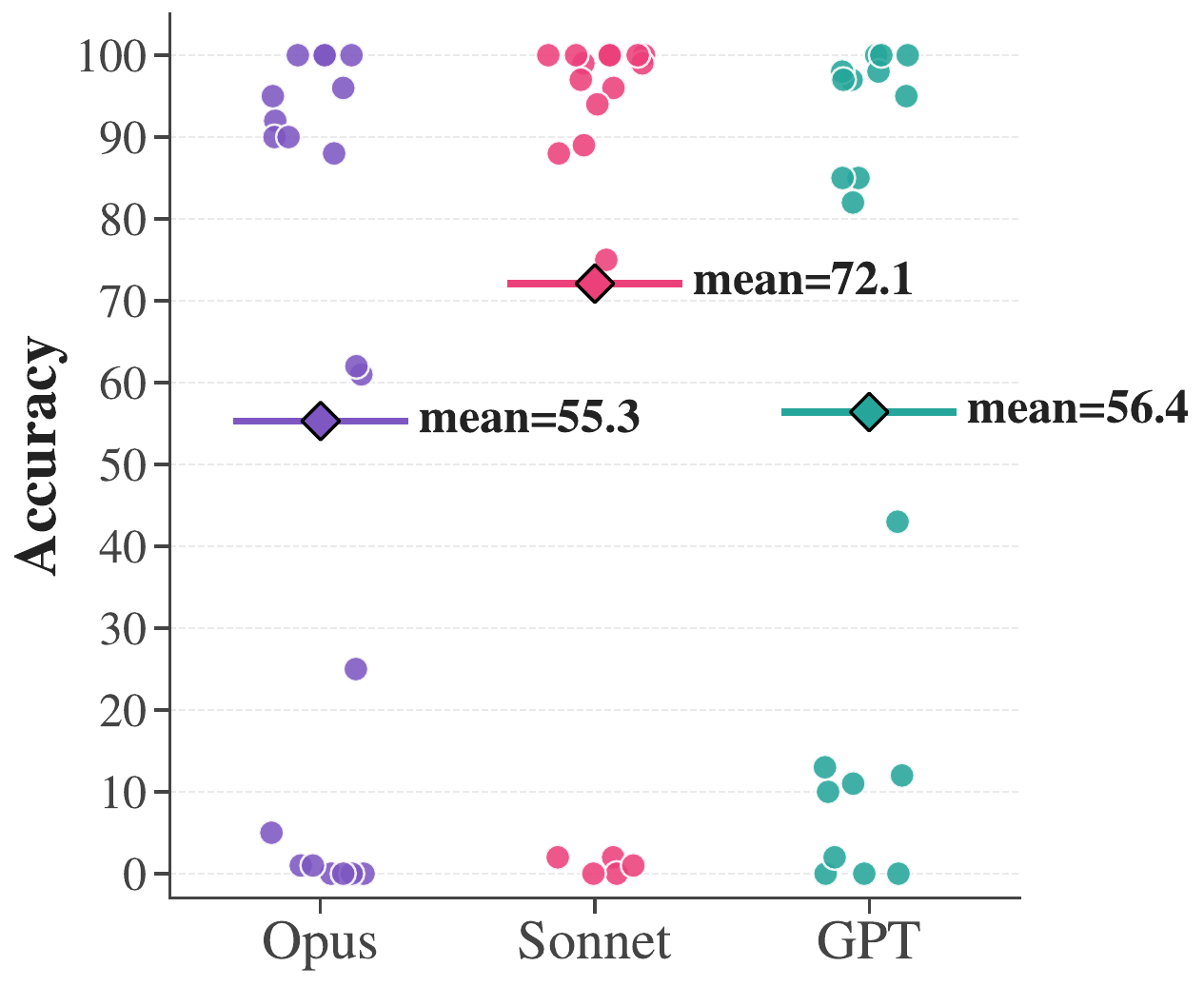}%
    }\hfill
  \subcaptionbox{AMC23\label{fig: amcinno}}{%
\includegraphics[width=0.31\linewidth]{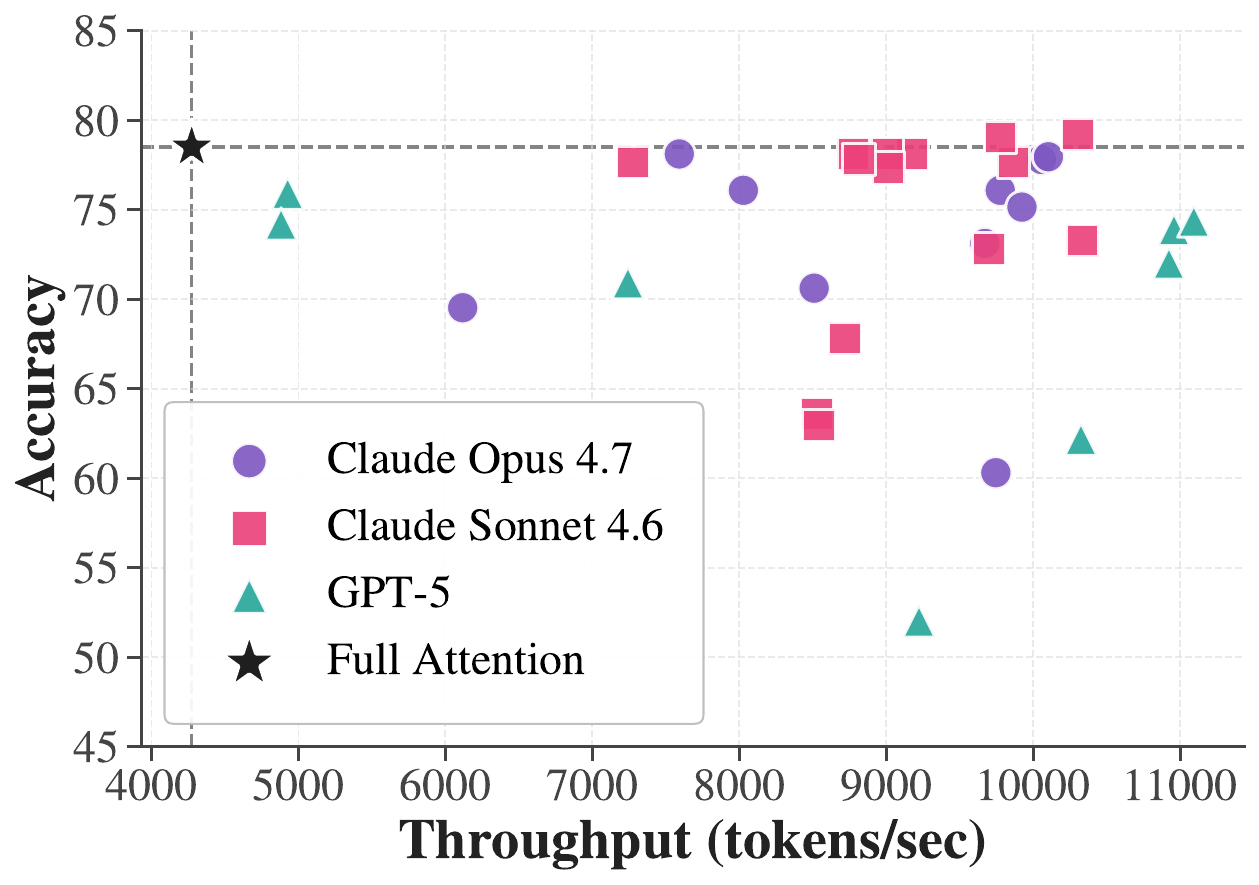}%
    }\hfill
    \subcaptionbox{AIME24\label{fig: aimeinno}}{%
\includegraphics[width=0.31\linewidth]{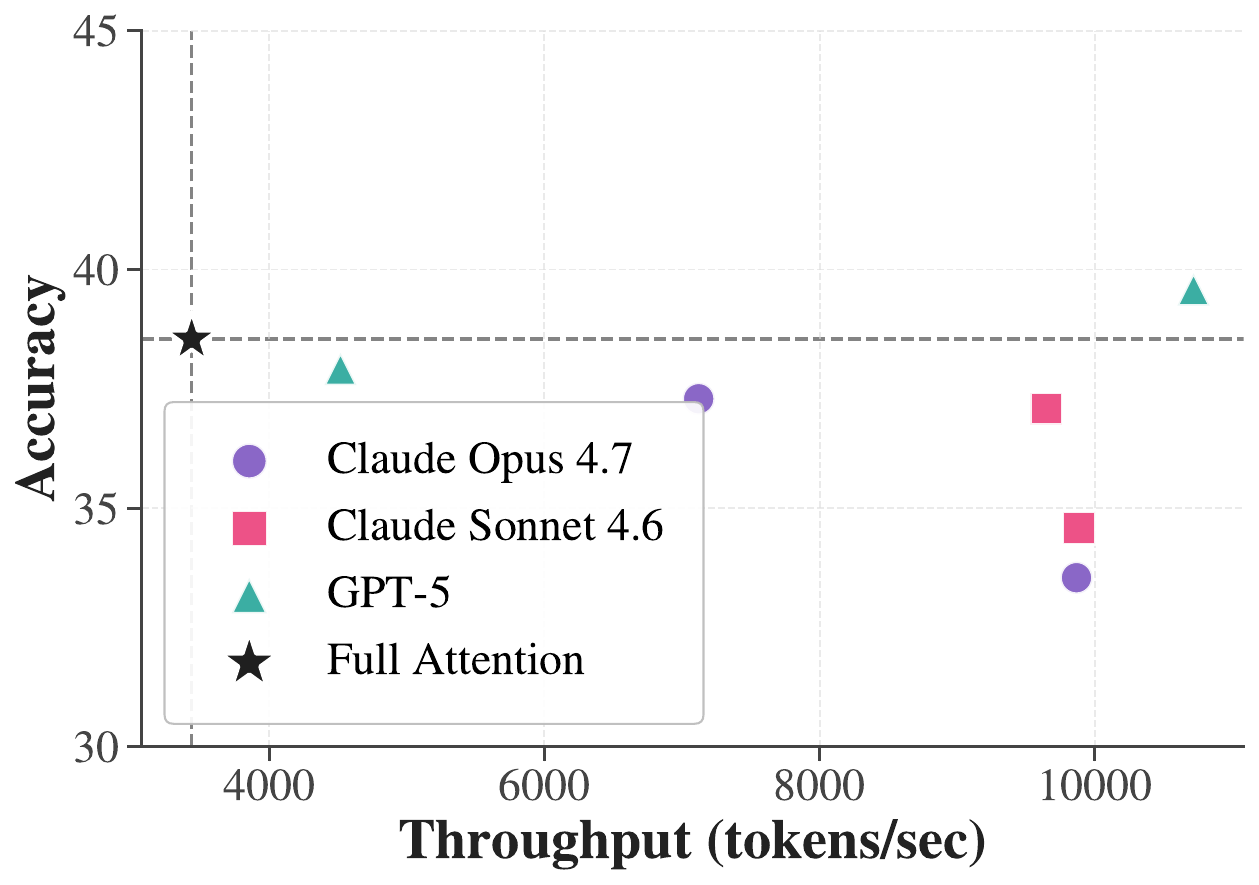}%
    }
  \caption{\textbf{Performance of AI-agent-generated sparse attention algorithms.} Each point corresponds to one generated algorithm evaluated by throughput and accuracy. AI agents consistently produce Pareto-efficient algorithms with full-attention accuracy and substantially higher throughput.}
\end{figure*}
\textbf{Algorithm Effectiveness.}
All generated algorithms compile and execute successfully in \sys without manual intervention, demonstrating the robustness and expressiveness of the programming model. We evaluate the generated algorithms using a staged pipeline: algorithms first run on RULER~\citep{hsieh2024ruler} 4K retrieval, those exceeding $85\%$ accuracy are evaluated on AMC23, and frontier algorithms are further tested on AIME24. As shown in~\Cref{fig: amcinno,fig: rulerinno,fig: aimeinno}, all three AI models produce Pareto-efficient sparse attention algorithms that maintain dense-attention-level accuracy while achieving $2\times$--$3.1\times$ higher throughput across benchmarks. Sonnet 4.6 produces the strongest algorithms on RULER and AMC23, while GPT-5 achieves the best AIME24 result. These results suggest that AI agents, when paired with \sys, can directly generate competitive sparse attention designs and accuracy--throughput trade-offs without iterative search or manual systems engineering.

\subsubsection{Experiment 2: Algorithm Iteration}
\label{sec:iteration}
The second study focuses on \emph{algorithm iteration}. The target benchmark is AIME24. We allow Claude Code to continuously modify, evaluate, and refine sparse attention algorithms over an 18-hour optimization loop. Unlike the previous study, this experiment does not require algorithmic novelty; the AI agent is free to optimize any aspect of the serving pipeline, including sparse attention structures, block sizes, KV-cache data types, and other system-level parameters. This experiment studies whether \sys can support rapid end-to-end experimentation and iterative improvement driven by AI agents.

\begin{figure*}
    \centering
    \subcaptionbox{Mean@16\label{fig: mean16iter}}{%
\includegraphics[width=0.31\linewidth]{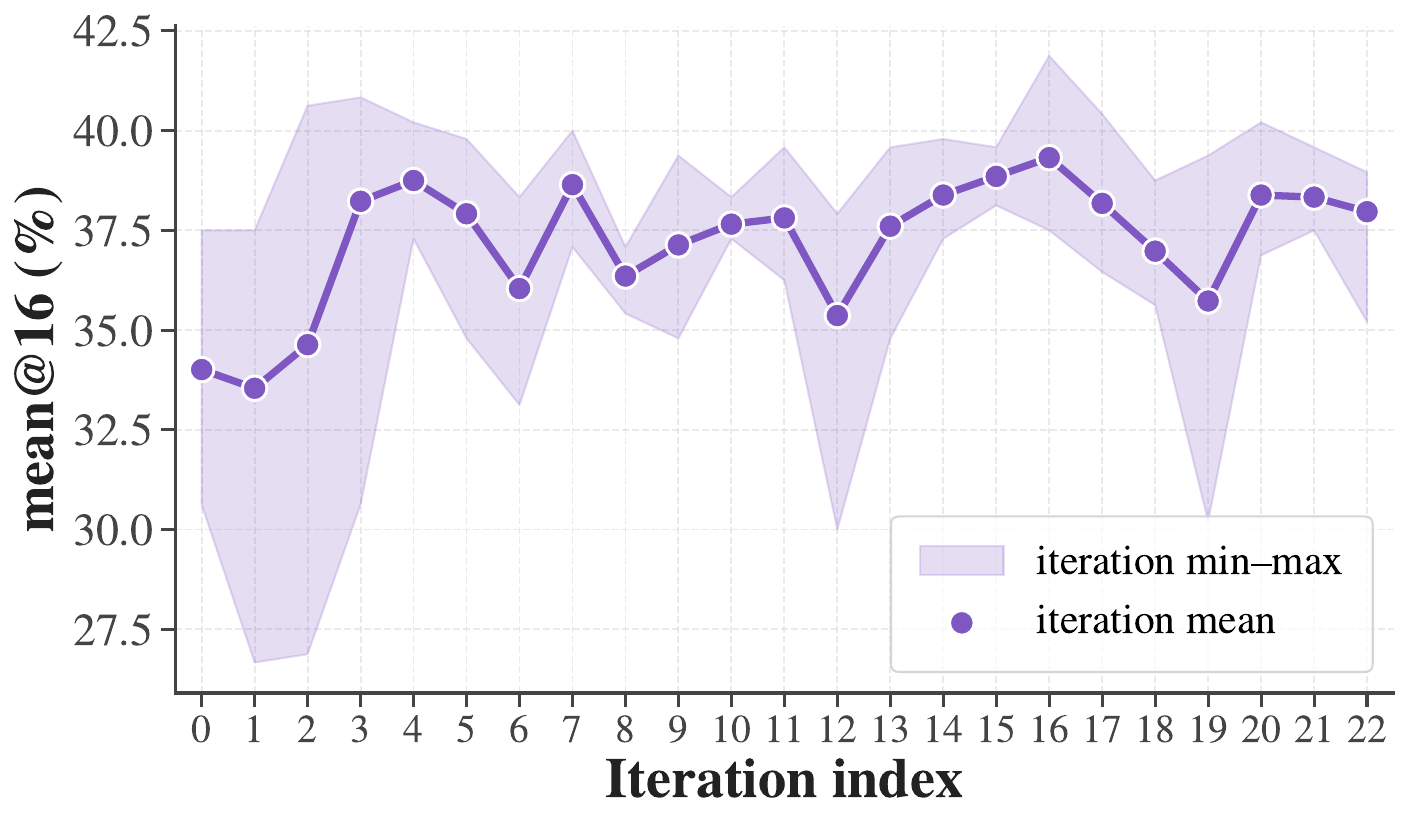}%
    }\hfill
  \subcaptionbox{Throughput\label{fig: throughput16iter}}{%
\includegraphics[width=0.31\linewidth]{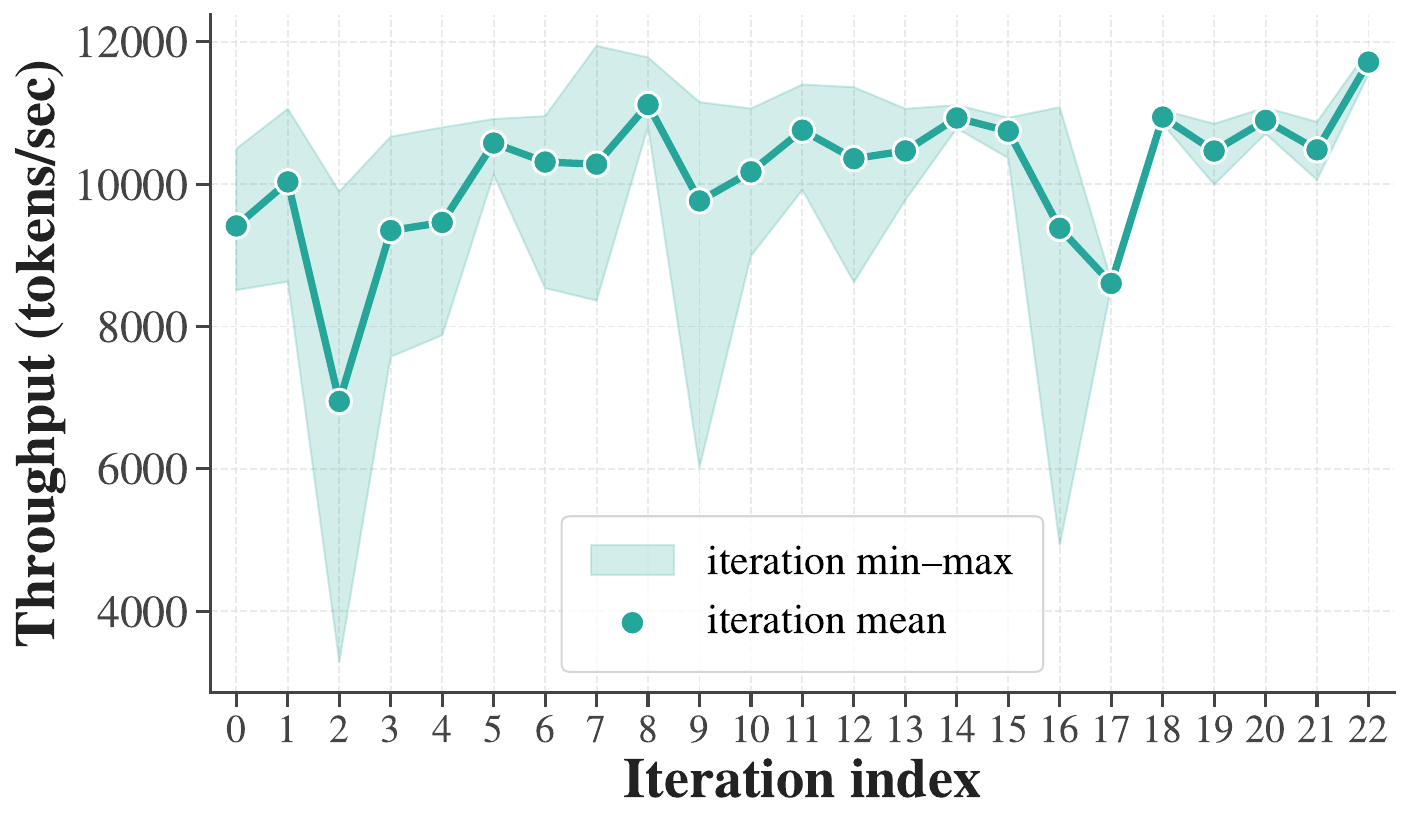}%
    }\hfill
    \subcaptionbox{Frontier\label{fig: iterationscatter}}{%
\includegraphics[width=0.31\linewidth]{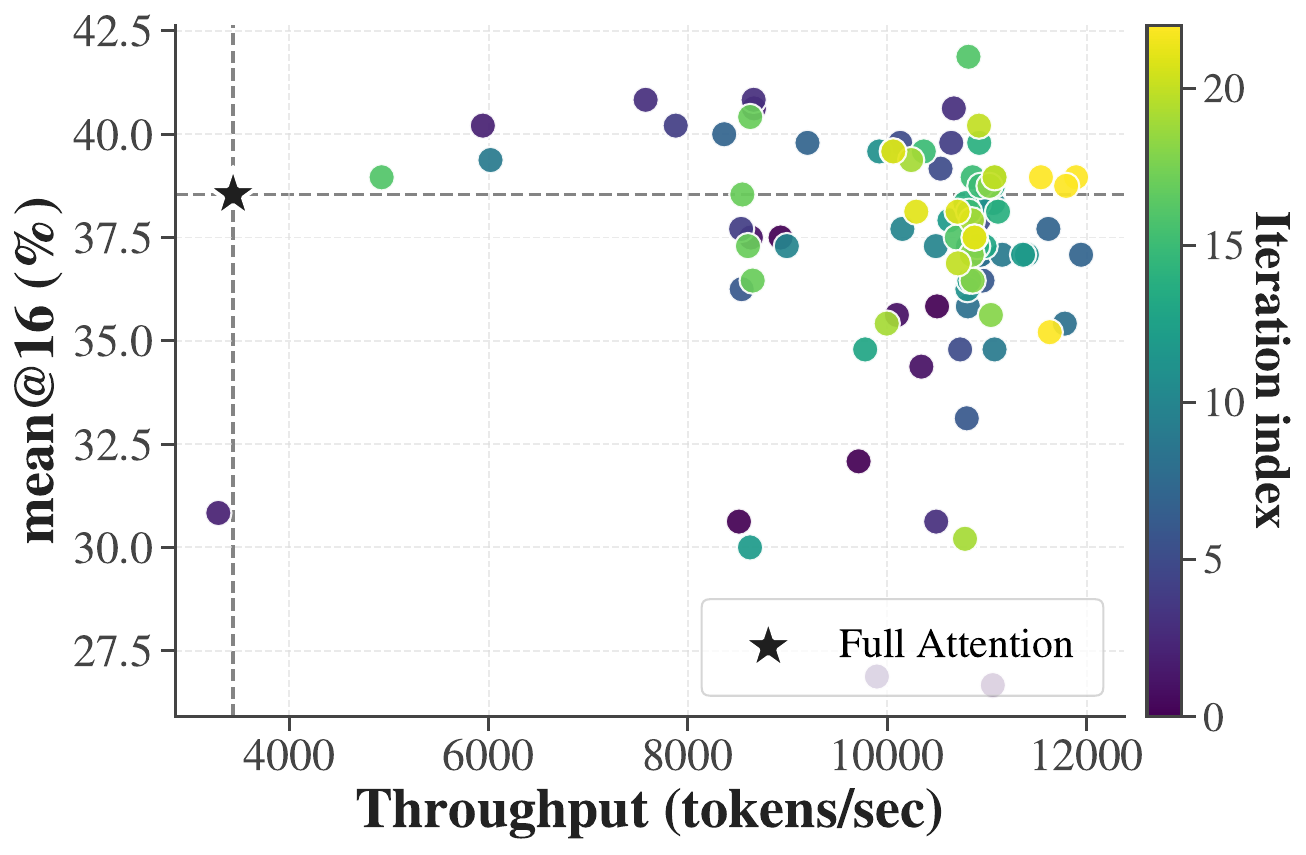}%
    }
  \caption{\textbf{Long-horizon autonomous optimization on AIME24} (Claude Opus 4.7, Qwen3-1.7B, 23 iterations, 92 total submissions). (a) Mean \texttt{mean@16} across variants per iteration. (b) Mean throughput (tokens/s) per iteration. Shaded regions show within-iteration min--max ranges. (c) Accuracy--throughput frontier of all submissions, colored by iteration order.}
\end{figure*}

We illustrate the results in~\Cref{fig: mean16iter,fig: throughput16iter,fig: iterationscatter}. The autonomous optimization loop produces sparse-attention flows that achieve up to a $3.46\times$ throughput speedup over dense attention on AIME24 ($11{,}894$ vs.\ $3{,}437$ tok/s) while preserving accuracy ($38.96$ vs.\ $38.54$).

Although the final algorithms converge to block top-$k$ attention rather than fundamentally new sparse attention mechanisms, the AI agent explores and evaluates many diverse algorithmic hypotheses throughout the optimization process, many of which outperform full attention. The improvements arise from a combination of algorithmic and systems-level refinements, including tuning block sizes, top-$k$ selection strategies, and skipped-layer configurations. More broadly, discovering fundamentally better sparse attention algorithms remains an open challenge, and we believe stronger experimental capabilities and longer optimization horizons could further improve AI-driven algorithm search


\subsection{Designing Sparse Attention on MLA Architectures}
\label{sec:mla}
\begin{figure*}
    \centering
    \subcaptionbox{mean@16\label{fig: glm-mean}}{%
\includegraphics[width=0.31\linewidth]{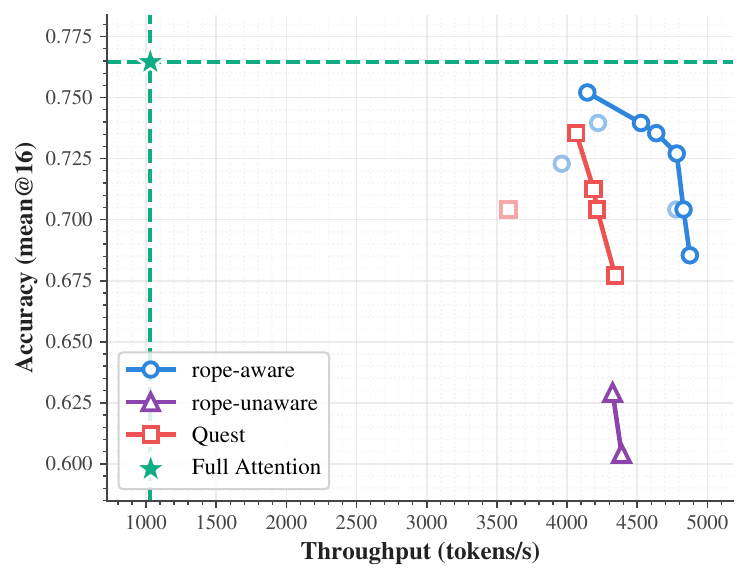}%
    }\hfill
  \subcaptionbox{pass@4\label{fig: glm-p4}}{%
\includegraphics[width=0.31\linewidth]{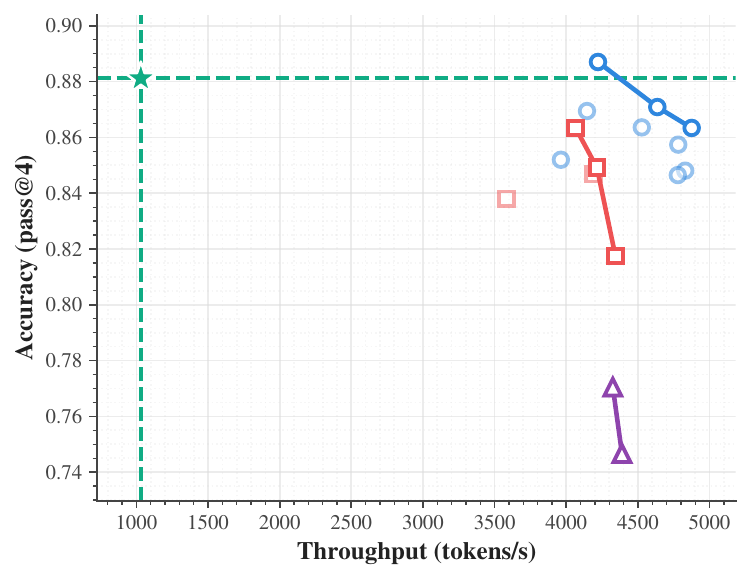}%
    }\hfill
    \subcaptionbox{pass@8\label{fig: glm-p8}}{%
\includegraphics[width=0.31\linewidth]{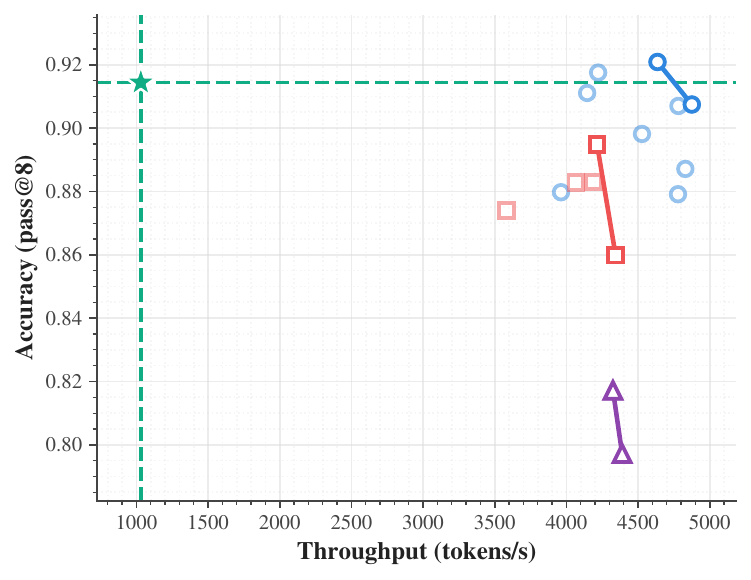}%
    }
  \caption{\textbf{Sparse attention on an MLA model.} Throughput versus accuracy for GLM-4.7-Flash on AIME26 with 32K-token generation on a single NVIDIA B200, reported as (a) mean@16, (b) pass@4, and (c) pass@8 against end-to-end throughput. Three MLA sparse-attention flows are expressed in \vflow (rope-aware block-sparse, rope-unaware block-sparse, and Quest), sweeping the number of attended blocks with block sizes of 16, 32, and 64.}
  \label{fig:glm-mla}
\end{figure*}

Beyond grouped-query attention, we study whether \sys generalizes to multi-head latent attention (MLA), the compressed-KV design used by DeepSeek~\citep{liu2025deepseek} and GLM~\citep{zeng2025glm}. This offers a second perspective on \textbf{Q1}. Instead of asking an agent to invent flows, we ask whether a researcher can design and validate architecture-specific sparse attention for a new attention family. We evaluate GLM-4.7-Flash on AIME26 with a 32K-token generation budget on a single NVIDIA B200 GPU,\footnote{GLM-4.7-Flash uses an MLA head geometry that the optimized vendor \texttt{trtllm\_mla} kernel does not support; dense attention therefore falls back to the much slower Triton MLA backend, which is why the full-attention baseline reaches only $1{,}031$ tokens/s, whereas \sys's sparse flows run on our optimized \texttt{cuda\_mla} backend.} designing a \emph{rope-aware} block-sparse flow in \vflow and comparing it against Quest. The rope-aware flow scores each KV block using the full fused-latent dot product, combining the compressed content and the decoupled rope components, a design specific to the MLA layout. (\Cref{fig:glm-mla} additionally includes a rope-unaware ablation, which we analyze in~\Cref{sec:understanding}.)

As shown in~\Cref{fig:glm-mla}, full attention reaches a mean@16 of $0.765$ but sustains only $1{,}031$ tokens/s. The rope-aware design matches this accuracy within about one point ($0.752$) while delivering roughly $4\times$ higher throughput, and up to $4.7\times$ with a tighter block budget, and it even matches or exceeds full attention on pass@4 and pass@8; Quest stays consistently behind the rope-aware frontier. This answers \textbf{Q1} from a second angle. With \sys, a researcher can rapidly design and validate sparse attention tailored to a new architecture, here turning the MLA layout into a $4\times$ end-to-end speedup at full-attention accuracy.

\subsection{Gaining New Understanding on Sparse Attention}
\label{sec:understanding}
\sys is also a research instrument. Because a scoring rule is only a few lines of \vflow, we can probe \emph{which} parts of the key representation actually drive block selection, turning sparse attention into a controlled experiment. We highlight two findings.

\begin{figure*}
    \centering
    \subcaptionbox{Qwen3-4B\label{fig: chan-loo4}}{%
\includegraphics[width=0.24\linewidth]{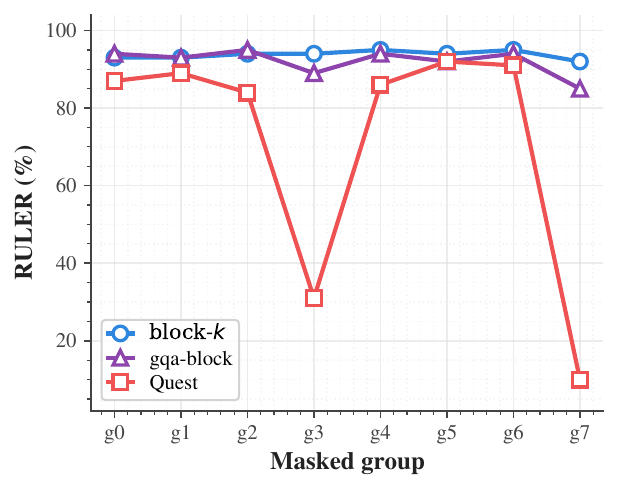}%
    }\hfill
    \subcaptionbox{Qwen3-8B\label{fig: chan-loo8}}{%
\includegraphics[width=0.24\linewidth]{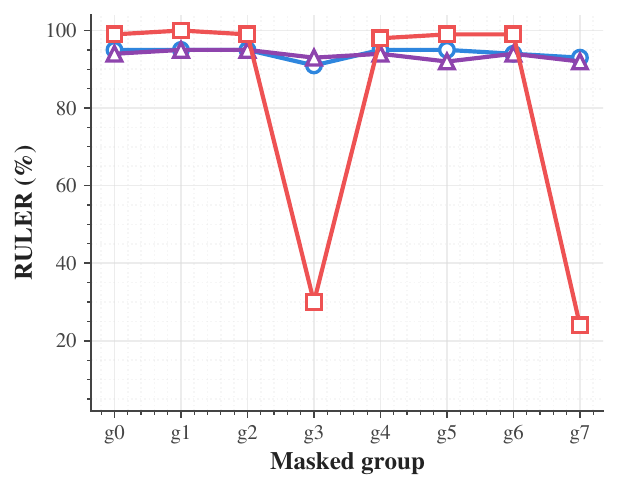}%
    }\hfill
    \subcaptionbox{Qwen3-4B\label{fig: chan-half4}}{%
\includegraphics[width=0.24\linewidth]{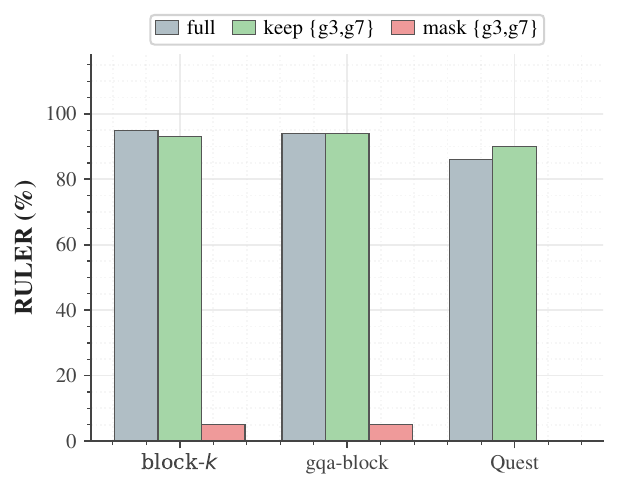}%
    }\hfill
    \subcaptionbox{Qwen3-8B\label{fig: chan-half8}}{%
\includegraphics[width=0.24\linewidth]{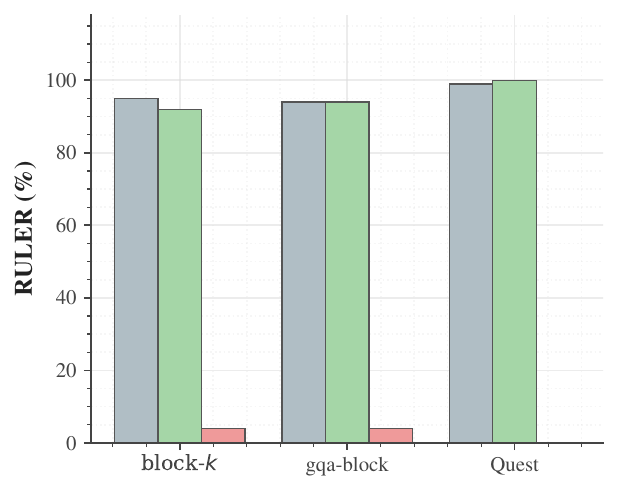}%
    }
  \caption{\textbf{Where the routing signal lives (RULER).} (a, b) Leave-one-out accuracy when masking one of the eight query-key channel groups g0 through g7. Only g3 and g7 matter, and Quest collapses when either is removed. (c, d) Keeping the four-group half that contains g3 and g7 is lossless, whereas masking g3 and g7 collapses every routing family to near-random. The same two groups are critical for both Qwen3-4B and Qwen3-8B.}
  \label{fig:channel}
\end{figure*}

\noindent\textbf{Query-key channel importance~\citep{Wang2026PrismSB}.} We split the $128$ query-key channels into eight groups of sixteen, labeled g0 through g7, and use \vflow to mask one group at a time during routing on Qwen3-4B and Qwen3-8B with RULER, then measure accuracy (\Cref{fig:channel}).\footnote{This channel-importance study is conducted on Qwen3 models (4B and 8B); the specific critical groups reflect the Qwen3 channel geometry and are not guaranteed to transfer to other model families.} Routing information turns out to be highly concentrated. Of the eight groups, only two (g3 and g7) are critical. Masking either one alone is nearly harmless, yet masking both collapses every method to near-random even when six of the eight groups are kept, so \emph{which} channels are kept matters far more than how many. Conversely, keeping just the half that contains g3 and g7 is lossless, and Quest even improves slightly. The same two groups are critical at both model sizes, indicating a stable Qwen3-family property of the channel geometry rather than a size-specific artifact.

\noindent\textbf{Rope component in MLA routing.} The MLA study in~\Cref{sec:mla} reveals an analogous effect along a different axis. The rope-unaware variant in~\Cref{fig:glm-mla}, which scores blocks from the compressed content alone and ignores the decoupled rope component, drops sharply to $0.60$ to $0.63$ mean@16, far below the rope-aware design at $0.75$. The positional rope component, not just the content, carries the signal needed for accurate block selection under MLA.

Together these studies expose a third facet of \textbf{Q1}. \sys is useful not only for building and deploying sparse attention but for understanding it, revealing where the routing signal lives across both GQA channels and MLA components.

\subsection{Efficiency Evaluations}
\label{sec:efficiency}
\noindent\textbf{Experimental Setup.}
We evaluate dense Qwen3 models~\citep{yang2025qwen3} ranging from 0.6B to 8B parameters on an NVIDIA H200 SXM GPU with 141GB memory. We study two sparse attention algorithms, block top-$k$ attention and Quest~\citep{tang2024quest}. Following prior work~\citep{zhang2023h2o,xiao2023efficient,chen2024magicpig}, each query always attends to the first block and the last two blocks, using a block size of 16 and varying the number of attended blocks from 32 to 256. We compare against SGLang v0.5.9~\citep{sglang}, into which \sys is integrated. To test scaling to larger models, longer generation, and fp8 precision, we additionally evaluate the 229B-parameter MiniMax-M2.7~\citep{chen2026minimax} across four NVIDIA B200 GPUs with tensor parallelism (TP$=4$); these scaling runs use up to 32K-token generation and sweep block sizes of 16, 32, and 64.


\subsubsection{Server-side Throughput}
\label{sec:end-to-end}
\begin{figure}
    \centering
    \subcaptionbox{Qwen3-0.6B\label{fig: 0.6bamc}}{%
\includegraphics[width=0.23\linewidth]{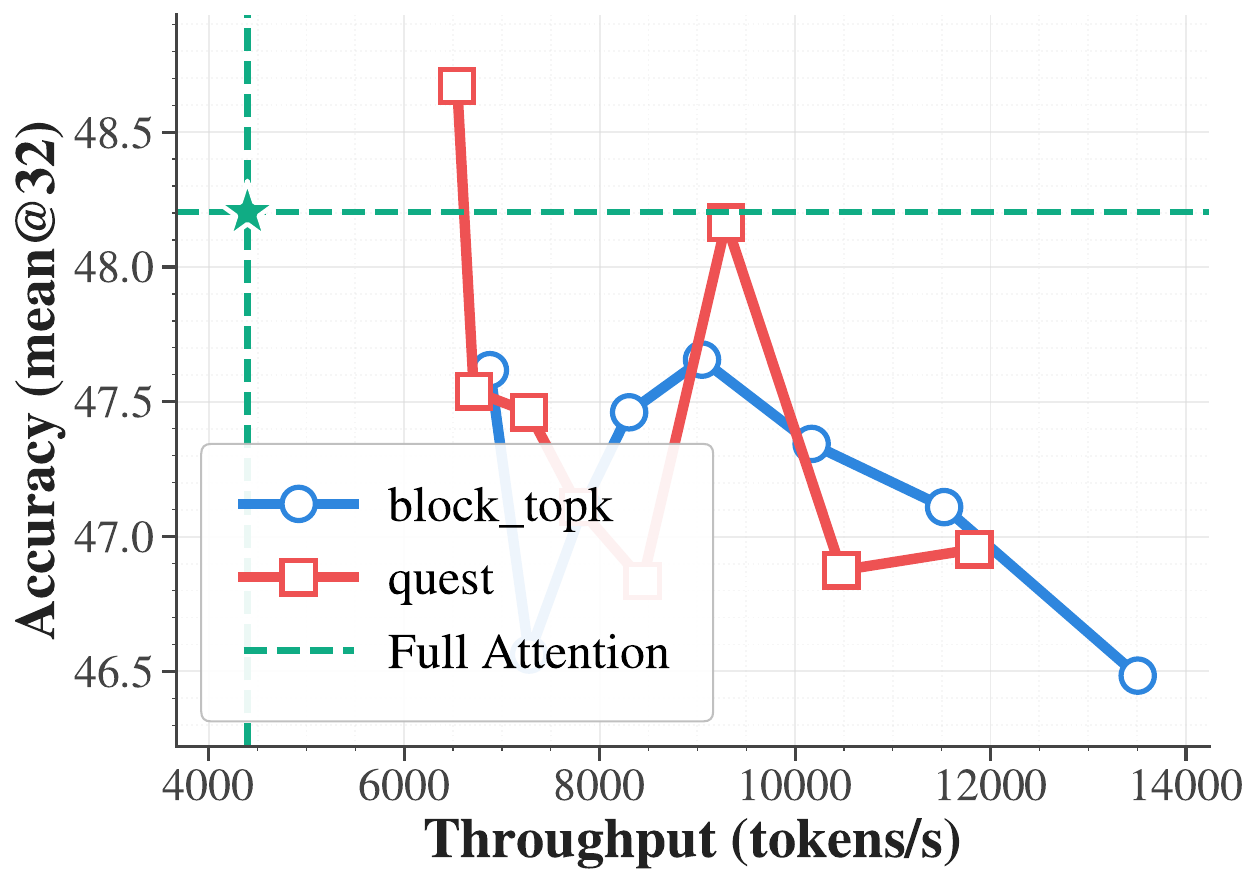}%
    }\hfill
  \subcaptionbox{Qwen3-1.7B\label{fig: 1.7bamc}}{%
\includegraphics[width=0.23\linewidth]{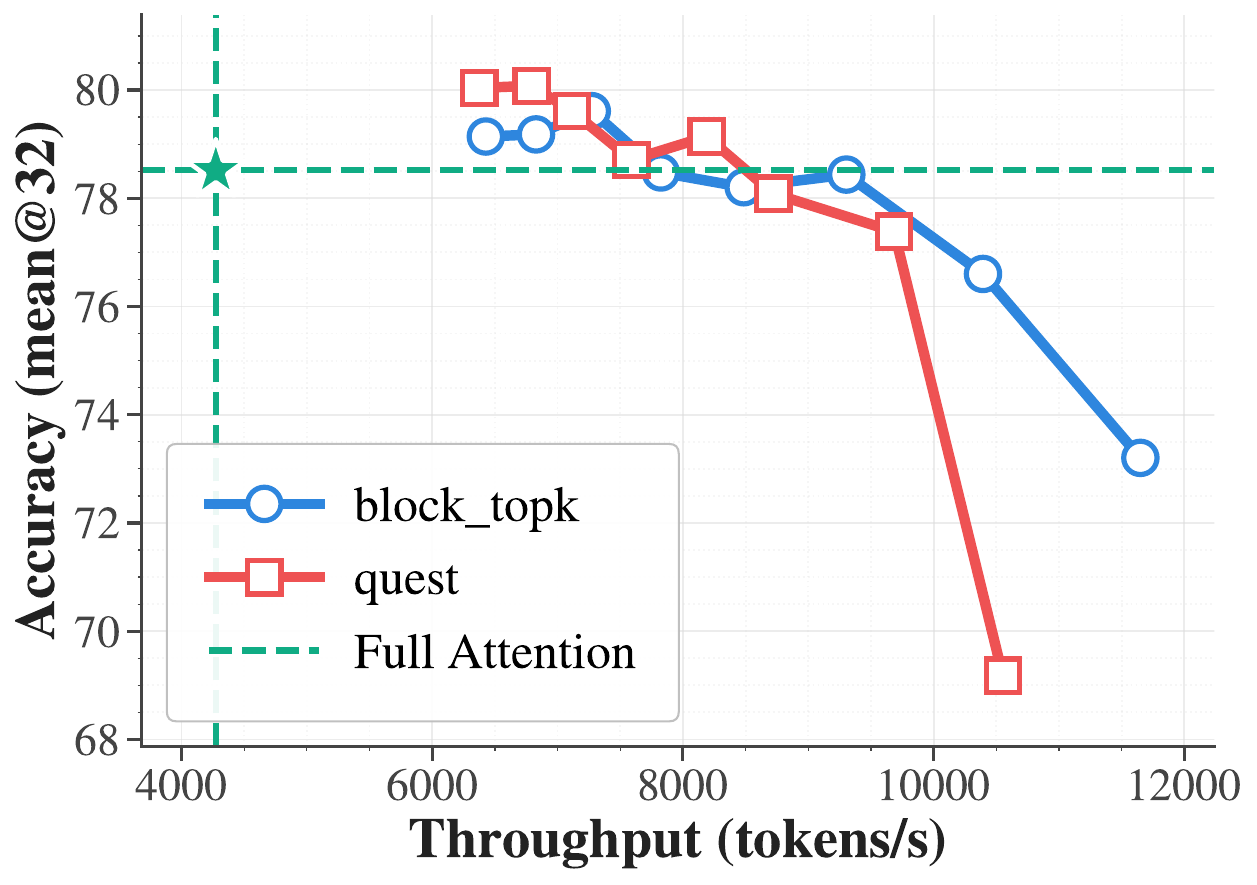}%
    }\hfill
    \subcaptionbox{Qwen3-4B\label{fig: 4bamc}}{%
\includegraphics[width=0.23\linewidth]{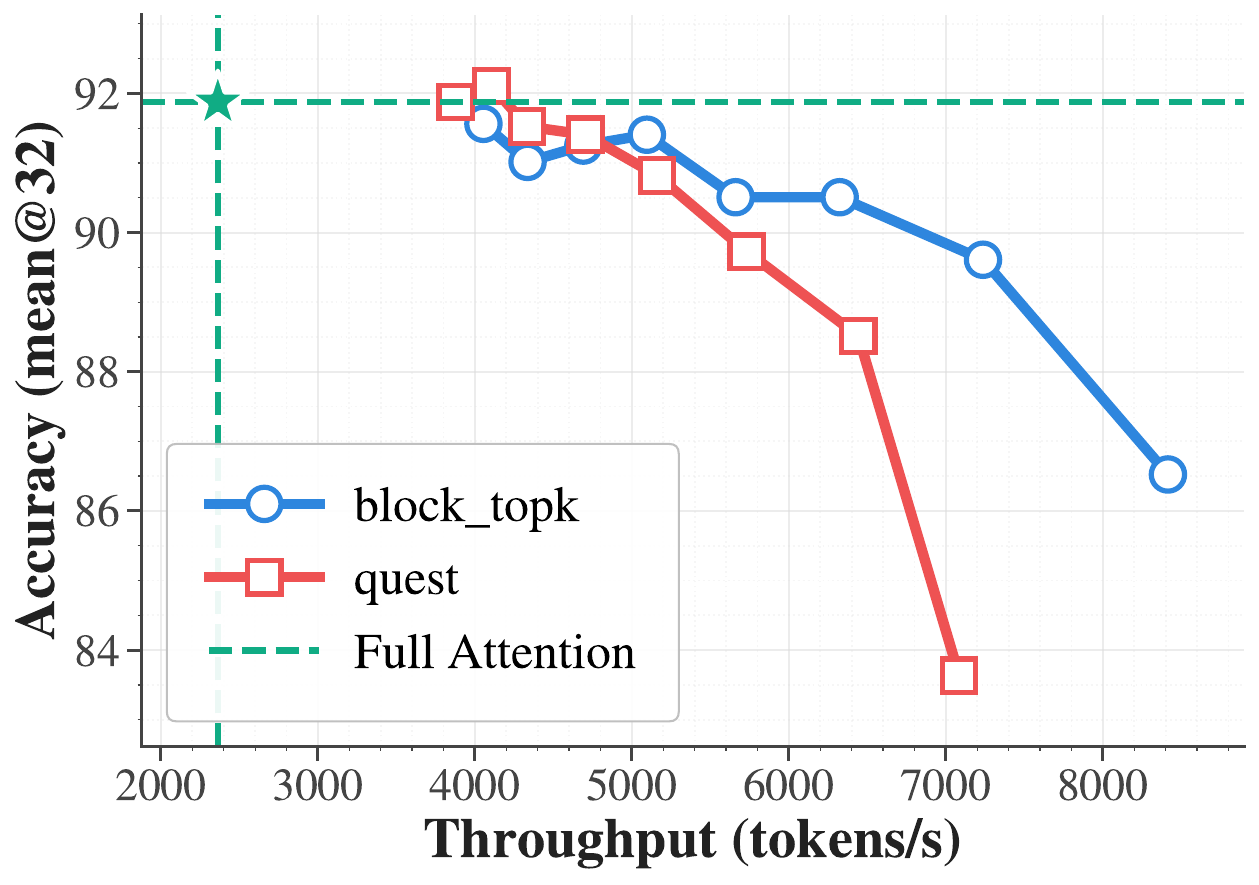}%
    }\hfill
    \subcaptionbox{Qwen3-8B\label{fig: 8bamc}}{%
\includegraphics[width=0.23\linewidth]{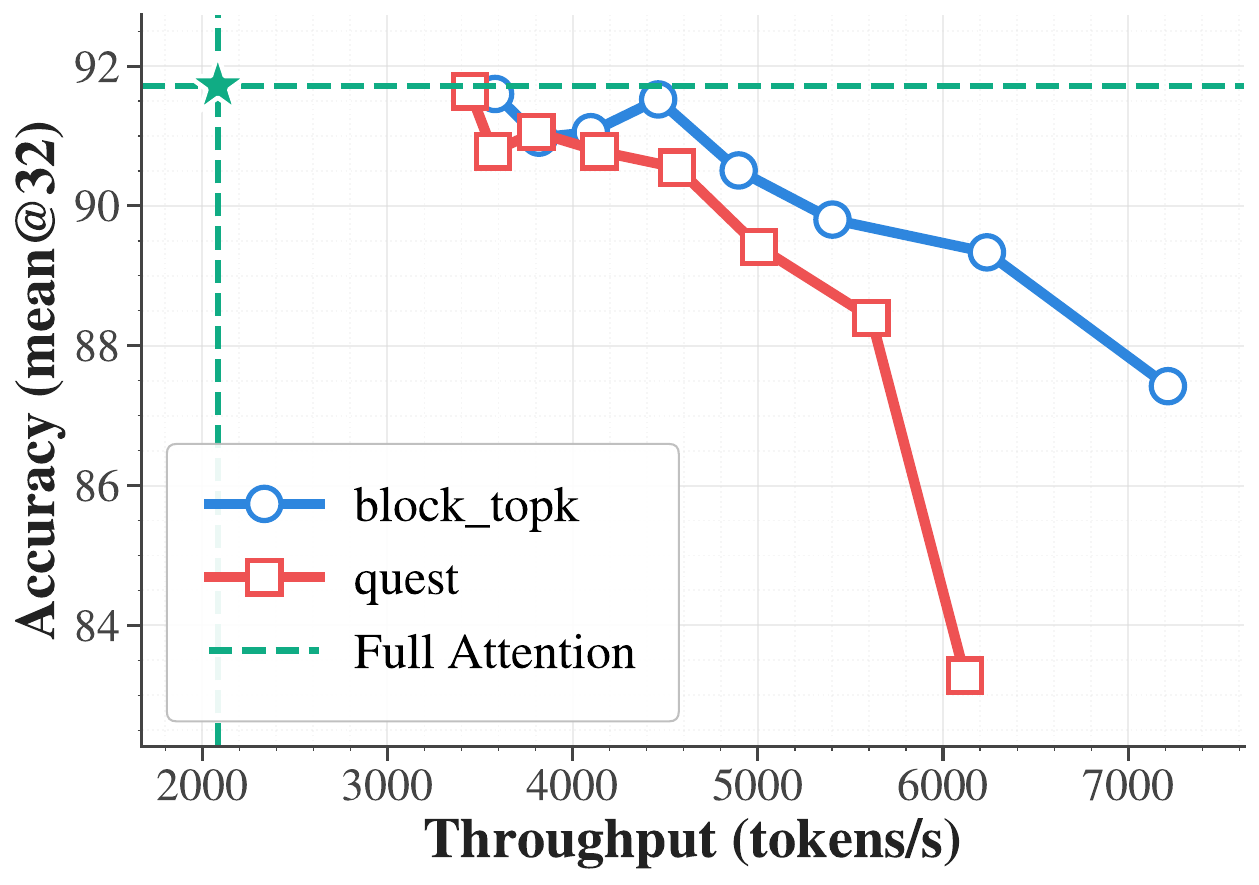}%
    } \\
    \subcaptionbox{Qwen3-0.6B\label{fig: 0.6baime}}{%
\includegraphics[width=0.23\linewidth]{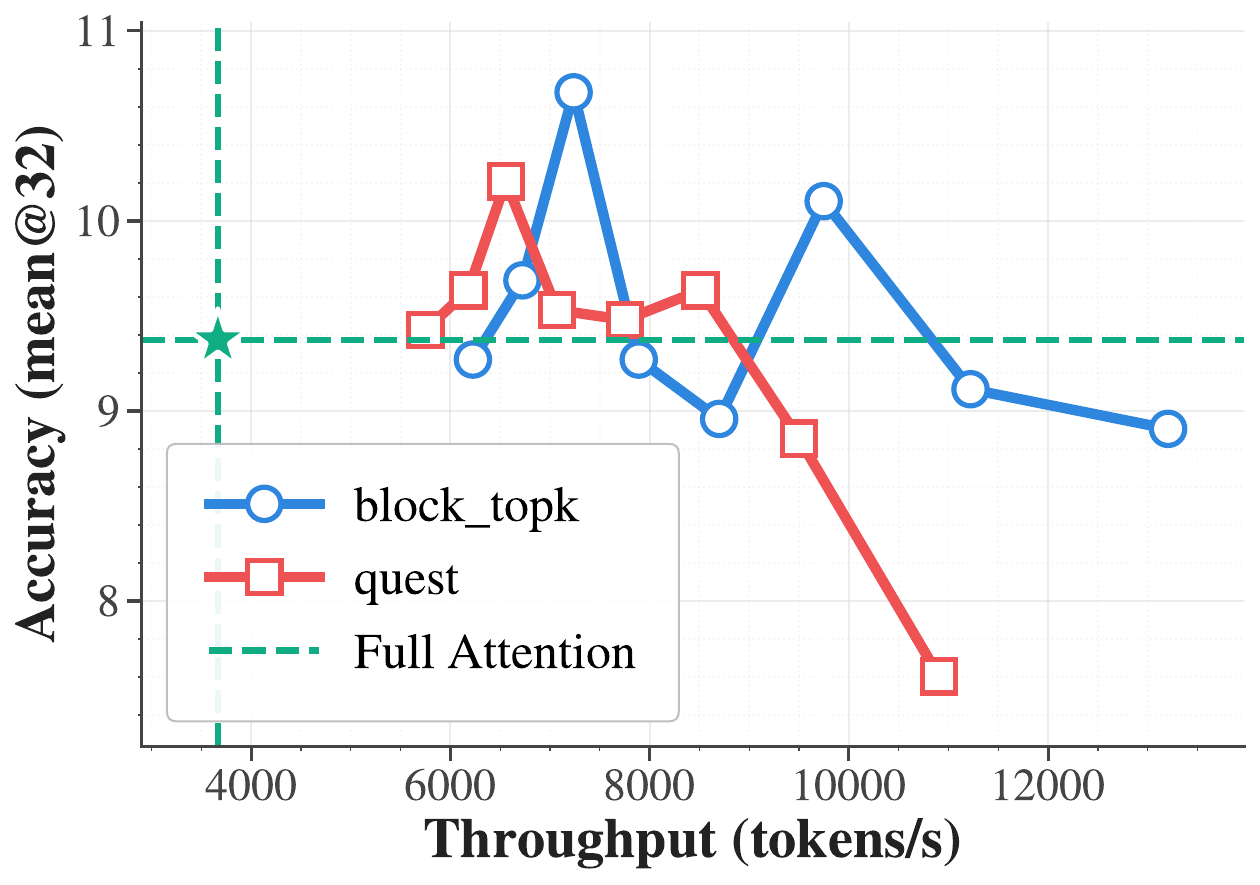}%
    }\hfill
  \subcaptionbox{Qwen3-1.7B\label{fig: 1.7baime}}{%
\includegraphics[width=0.23\linewidth]{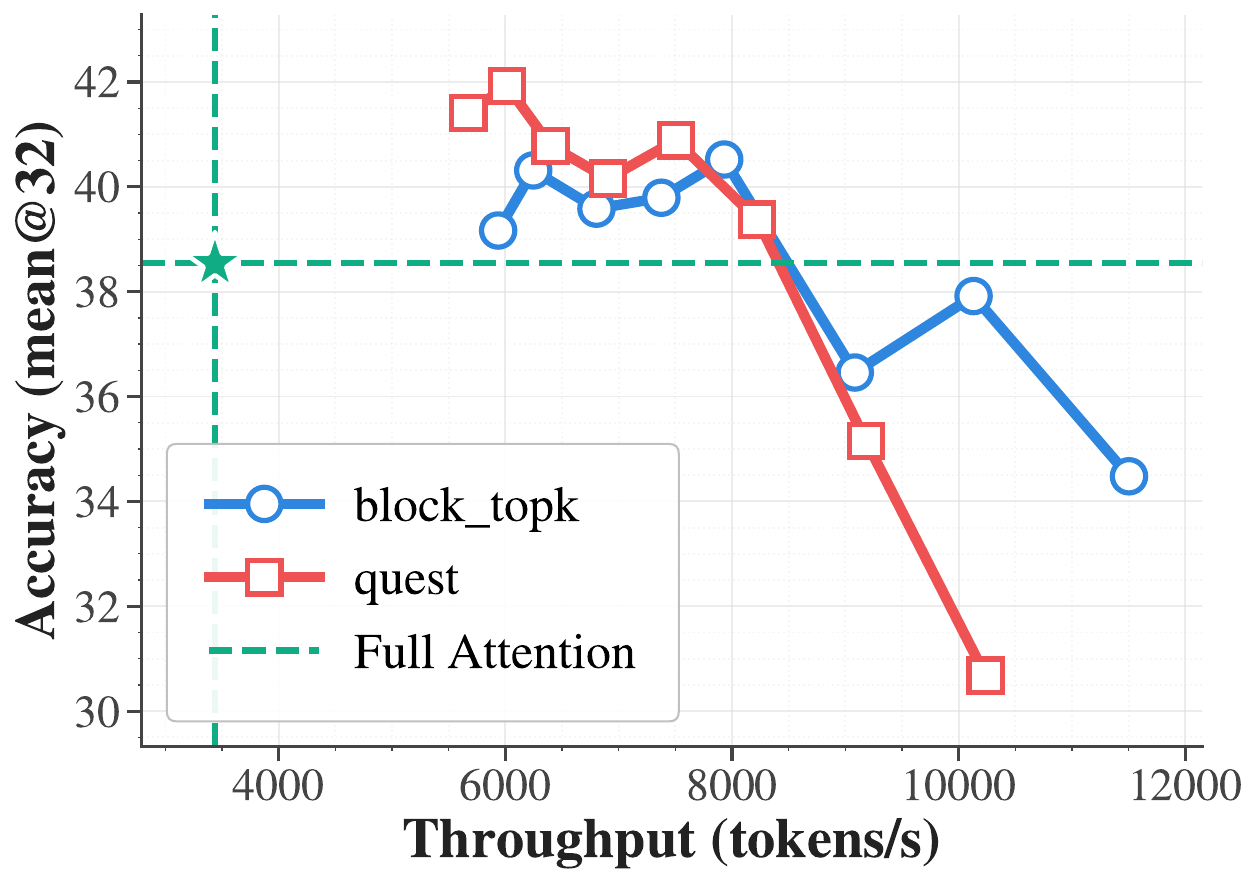}%
    }\hfill
    \subcaptionbox{Qwen3-4B\label{fig: 4baime}}{%
\includegraphics[width=0.23\linewidth]{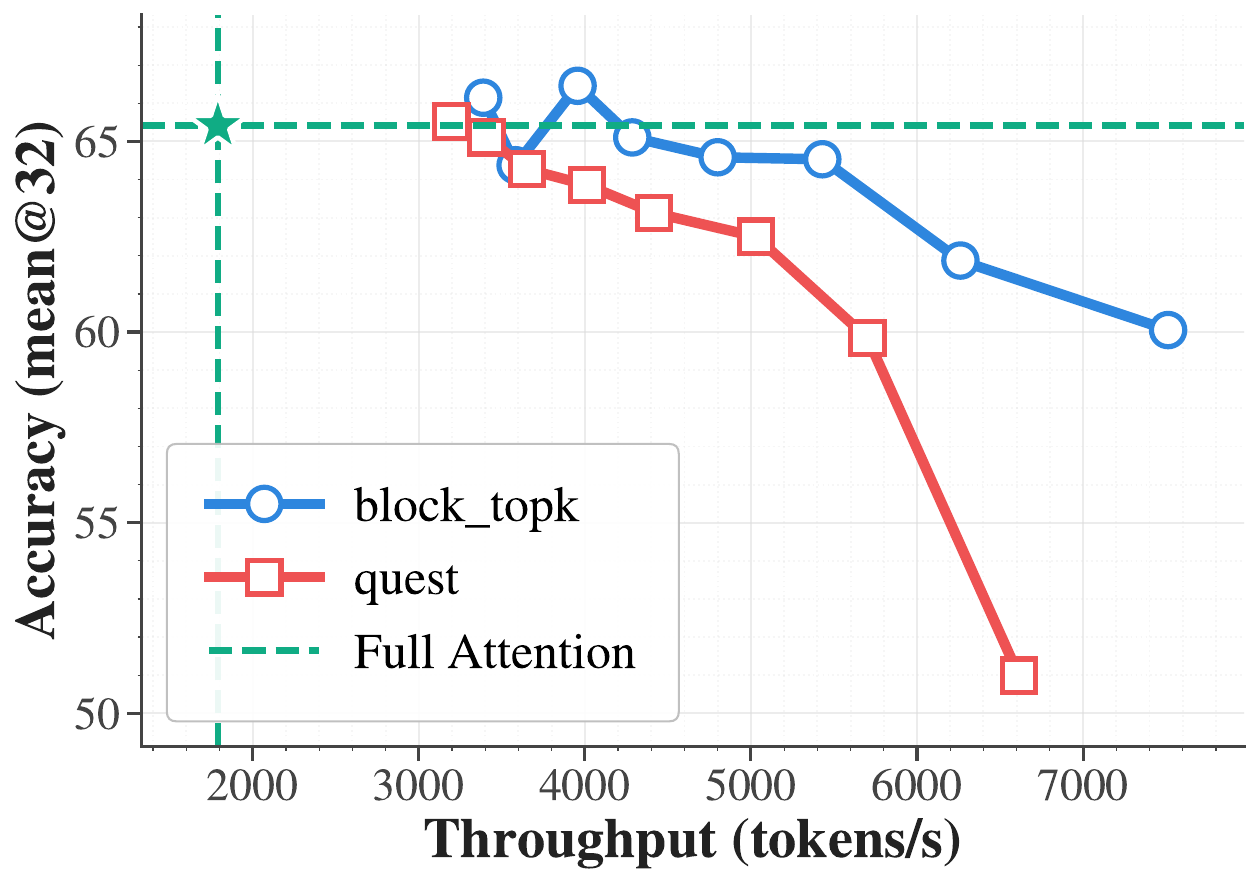}%
    }\hfill
    \subcaptionbox{Qwen3-8B\label{fig: 8baime}}{%
\includegraphics[width=0.23\linewidth]{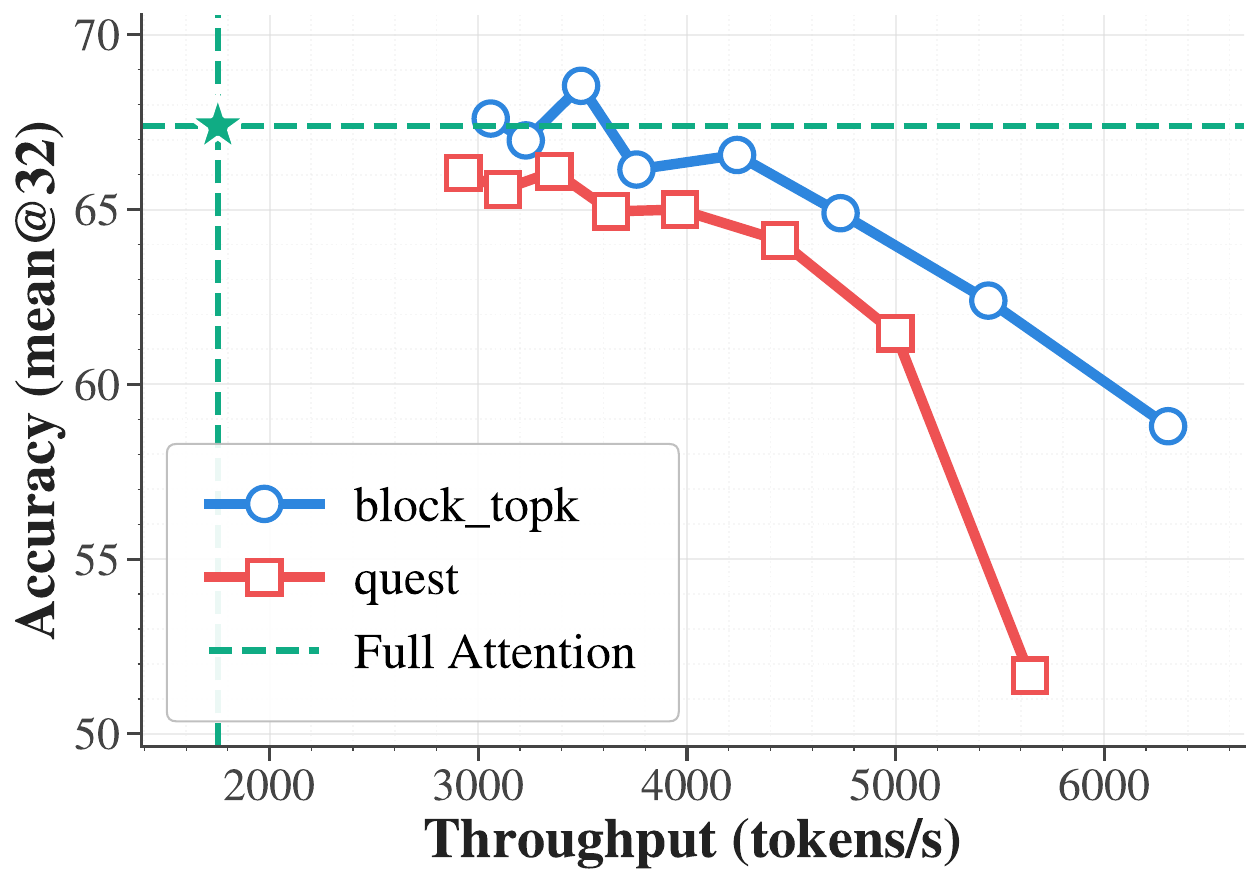}%
    }
   \caption{\textbf{Throughput--accuracy Pareto frontiers on AMC23 (a--d) and AIME24 (e--h)} for Qwen3-0.6B, 1.7B, 4B, and 8B as the number of attended blocks is swept from 32 to 256. The dashed line and star indicate the operating point of full attention.}
\end{figure}


\noindent\textbf{Dense models.}
We evaluate server-side throughput on the long-generation benchmarks AMC23 and AIME24, with input lengths up to 4K tokens and outputs up to 16K tokens. We report end-to-end offline generation throughput and construct throughput--accuracy Pareto frontiers. As shown in~\Cref{fig: 0.6baime,fig: 0.6bamc,fig: 1.7baime,fig: 1.7bamc,fig: 4baime,fig: 4bamc,fig: 8baime,fig: 8bamc}, sparse attention achieves substantial throughput gains. Under a 5 pp accuracy budget, block top-$k$ attains up to $3.46\times$ and $3.60\times$ speedup on AMC23 and AIME24, respectively, while Quest achieves up to $2.73\times$ and $2.98\times$. Even under a stricter 1 pp accuracy budget, block top-$k$ still delivers $2.14$--$2.31\times$ speedup on AMC23 and $2.42$--$3.60\times$ on AIME24, while Quest achieves $1.99$--$2.11\times$ and $1.90$--$2.59\times$, respectively.

\begin{figure*}
    \centering
    \subcaptionbox{mean@16\label{fig: mm-mean}}{%
\includegraphics[width=0.31\linewidth]{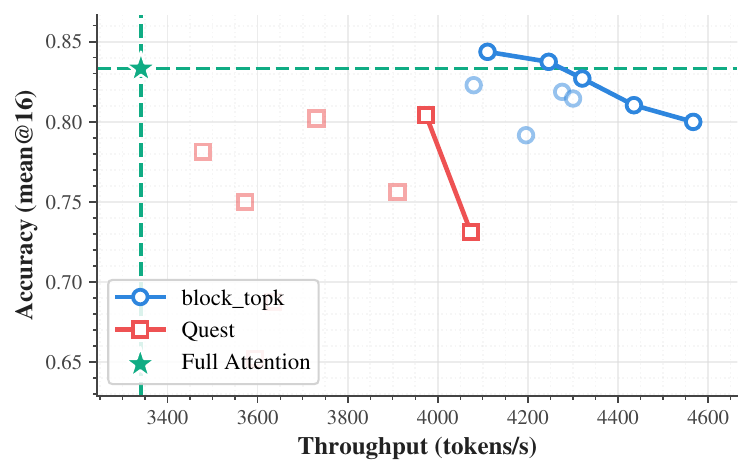}%
    }\hfill
  \subcaptionbox{pass@4\label{fig: mm-p4}}{%
\includegraphics[width=0.31\linewidth]{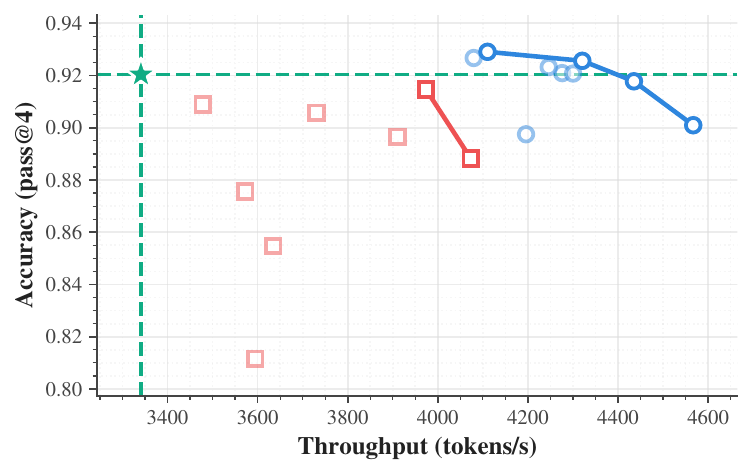}%
    }\hfill
    \subcaptionbox{pass@8\label{fig: mm-p8}}{%
\includegraphics[width=0.31\linewidth]{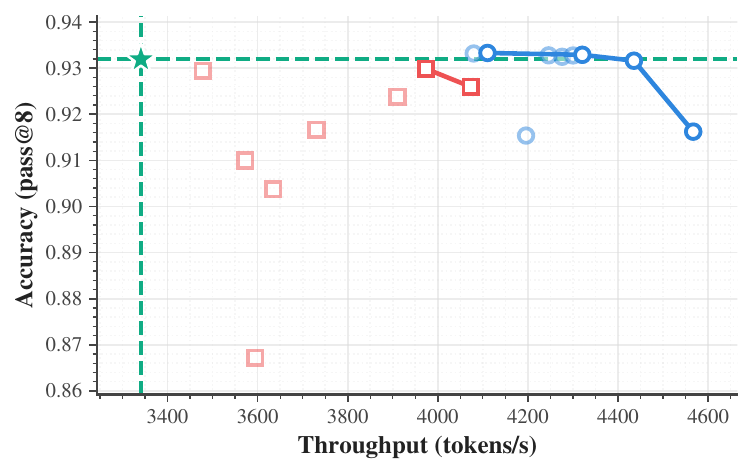}%
    }
  \caption{\textbf{Scaling to a 229B model with tensor parallelism.} Throughput versus accuracy for MiniMax-M2.7 (229B) on AIME26 with 32K-token generation on four NVIDIA B200 GPUs (TP$=4$), reported as (a) mean@16, (b) pass@4, and (c) pass@8 against end-to-end throughput. Block top-$k$ and Quest sweep the number of attended blocks; the star marks the full-attention operating point.}
  \label{fig:minimax}
\end{figure*}

\noindent\textbf{Scaling to a 229B MoE with tensor parallelism.}
We push further to MiniMax-M2.7, a 229B-parameter model served across four NVIDIA B200 GPUs with tensor parallelism (TP$=4$), again on AIME26 with 32K-token generation (\Cref{fig:minimax}). Full attention reaches a mean@16 of $0.83$ at $3{,}341$ tokens/s. Block top-$k$ even slightly exceeds this accuracy ($0.84$) while running $1.23\times$ faster ($4{,}110$ tokens/s), and reaches up to $1.37\times$ with a tighter block budget, while Quest attains $1.19\times$ at comparable accuracy. The relative gains are smaller than for the dense and 30B-MoE models because at this scale, with TP$=4$, a larger share of decode time goes to parameter and communication traffic rather than KV-cache access, yet sparse attention still delivers a consistent end-to-end speedup at the largest scale we test.

\noindent\textbf{Choosing block size and budget.}
Sweeping the block size (16, 32, 64) and the top-$k$ budget across these experiments, we observe a consistent pattern. A small block size of 16 with a moderate budget of around 125 attended blocks sits near the accuracy ceiling while retaining most of the throughput gain, larger blocks favor throughput, and raising the budget further yields diminishing accuracy returns; block-16 with top-$k\approx125$ is thus a robust default across model scales. \sys made these sweeps, spanning 0.6B to 229B models, practical to run fully end-to-end.

\subsubsection{User-side Latency}
\label{sec:latency}
\begin{figure*}
    \centering
    \subcaptionbox{Qwen3-0.6B\label{fig: 0.6b16k}}{%
\includegraphics[width=0.23\linewidth]{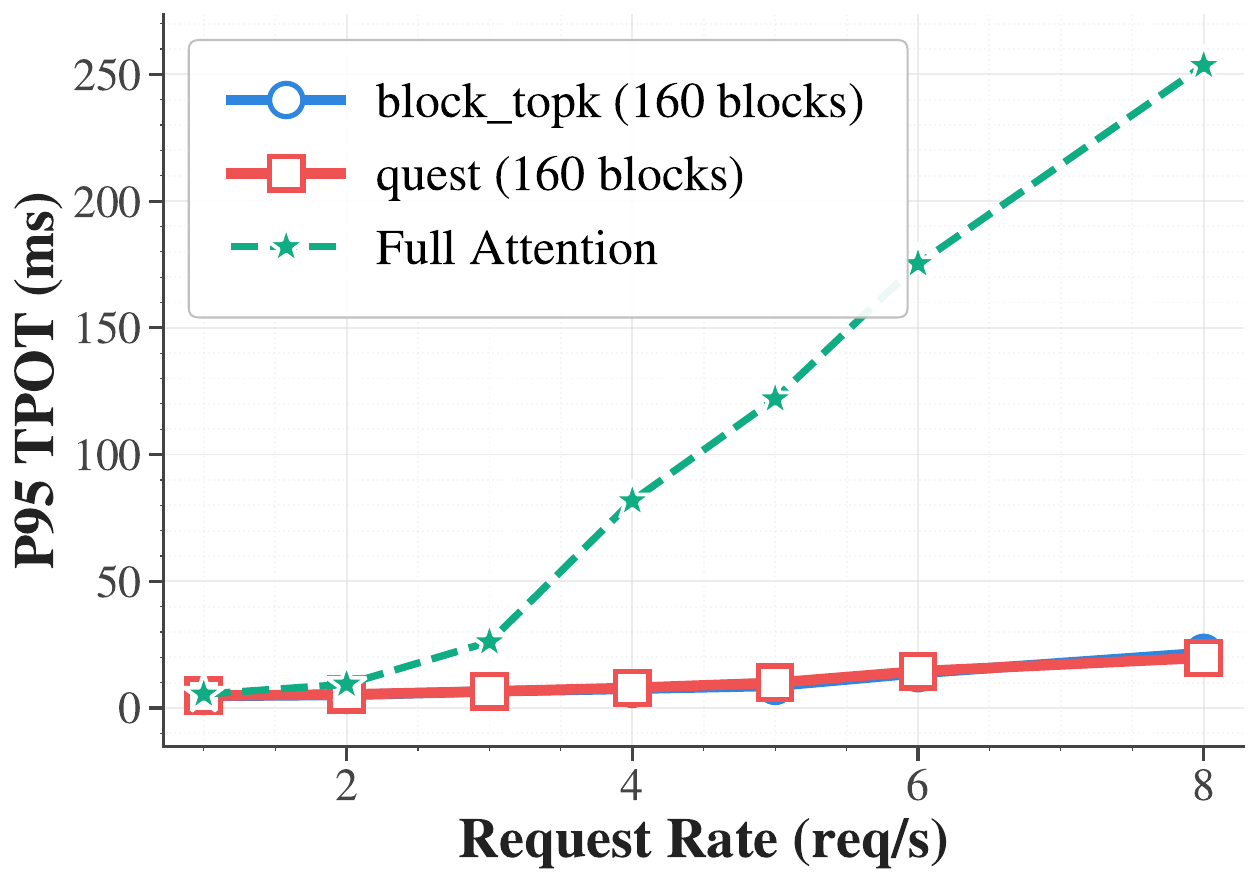}%
    }\hfill
  \subcaptionbox{Qwen3-1.7B\label{fig: 1.7b16k}}{%
\includegraphics[width=0.23\linewidth]{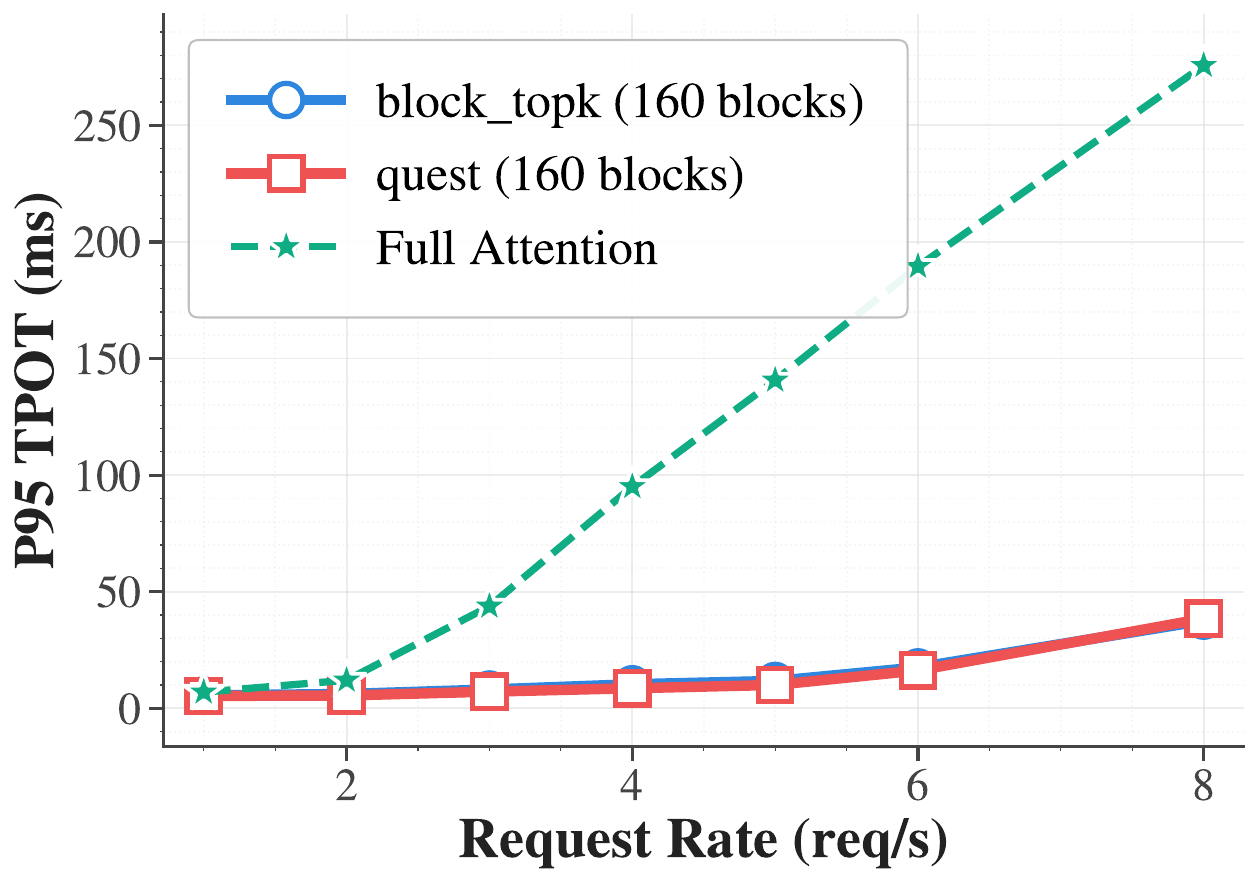}%
    }\hfill
    \subcaptionbox{Qwen3-4B\label{fig: 4b16k}}{%
\includegraphics[width=0.23\linewidth]{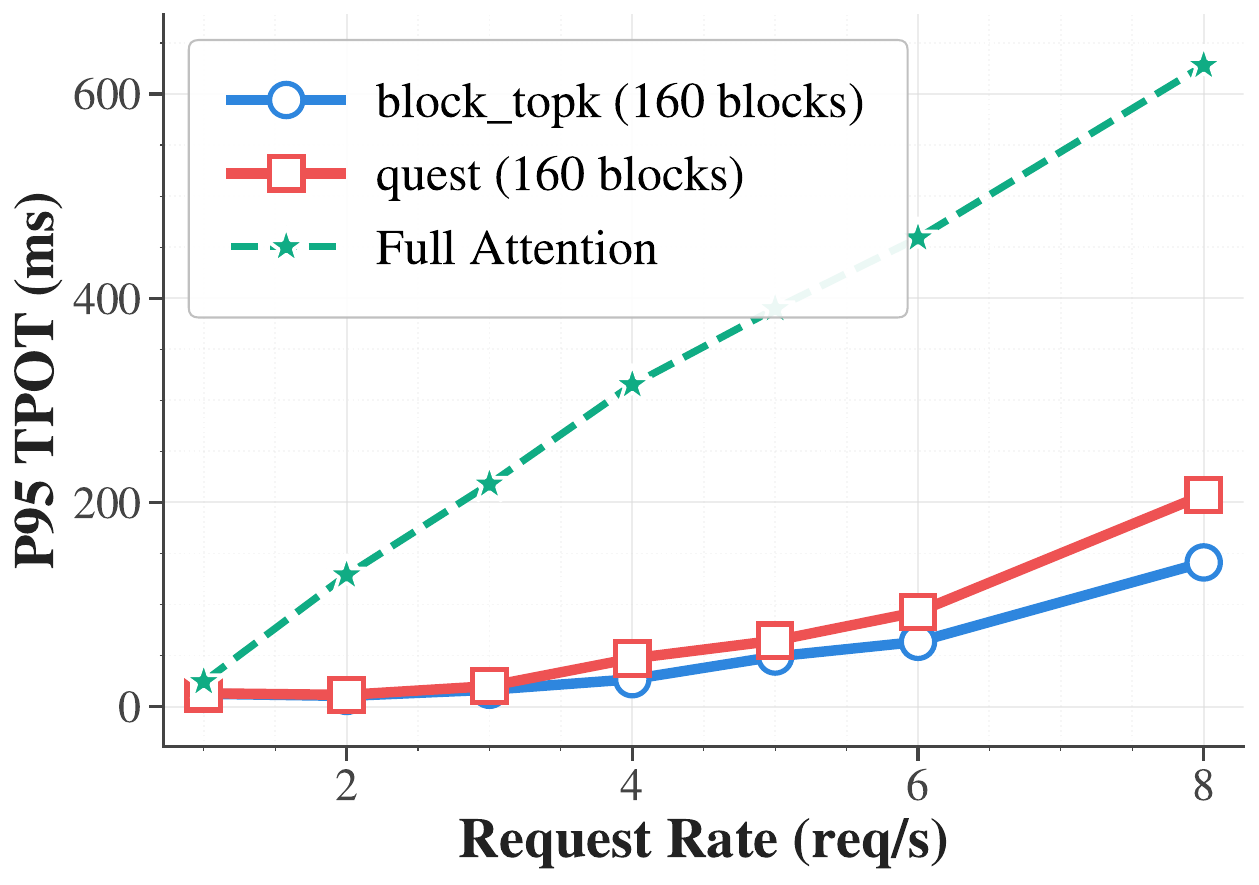}%
    }\hfill
    \subcaptionbox{Qwen3-8B\label{fig: 8b16k}}{%
\includegraphics[width=0.23\linewidth]{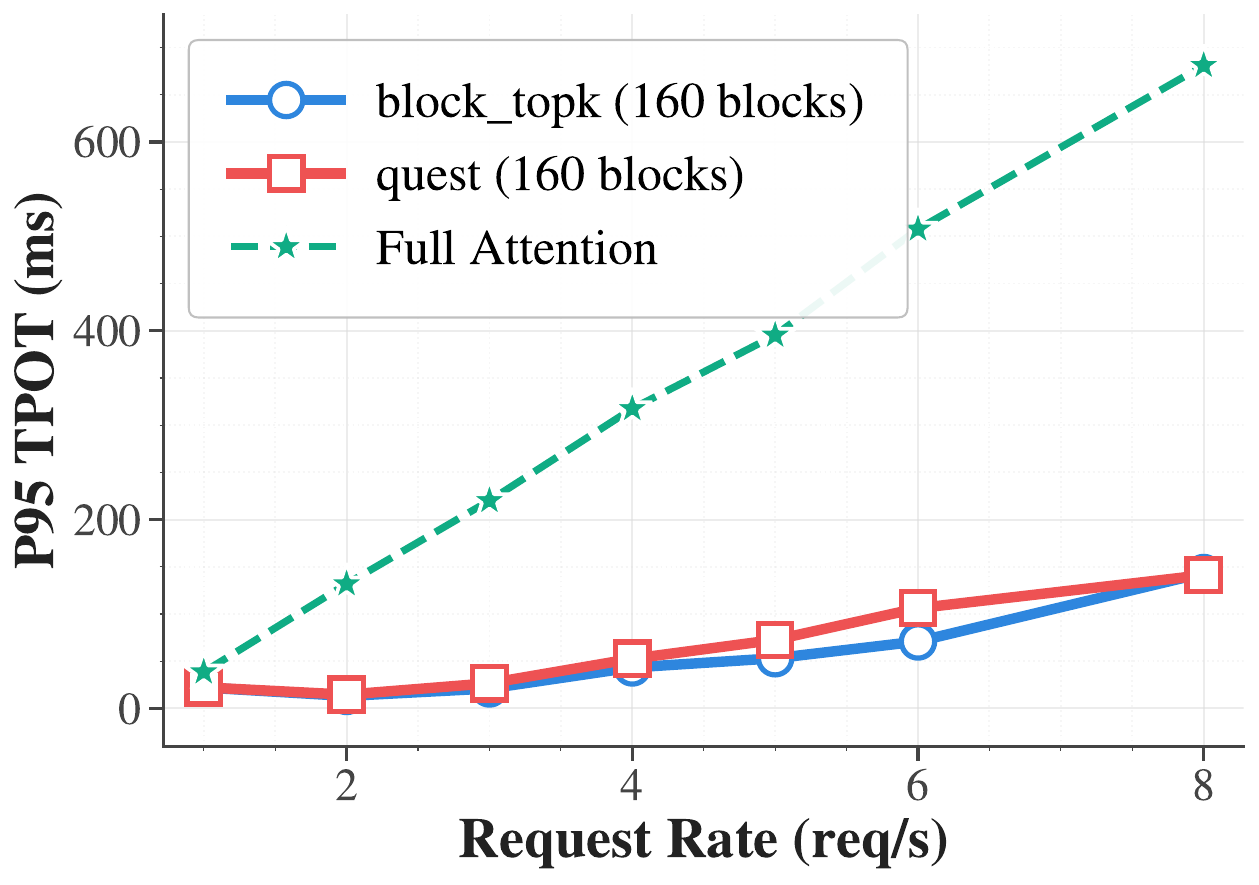}%
    }
   \caption{\textbf{User-side latency at 16K input} for Qwen3 series. We plot P95 TPOT versus request rate for full attention, block top-$k$, and quest at 160 attended blocks.}
\end{figure*}

We evaluate user-side latency using synthetic 16K-token prompts at request rates from 1.0 to 8.0 req/s, with each request generating 512 tokens. We report P95 time per output token (TPOT). All experiments use a P1D1-disaggregated setup to isolate decoding performance, the primary target of \sys's sparse-attention optimizations. As shown in~\Cref{fig: 0.6b16k,fig: 1.7b16k,fig: 4b16k,fig: 8b16k}, sparse attention substantially reduces latency under high request rates, achieving up to $11.7\times$ lower P95 TPOT with block top-$k$ and $12.8\times$ lower P95 TPOT with Quest at 8 req/s. At moderate load (4 req/s), block top-$k$ reduces latency by up to $11.8\times$, while Quest achieves up to $11.1\times$ reduction. 

\section{Conclusion}
We presented \textsc{Vortex}, a system for rapidly developing and deploying sparse-attention algorithms in modern LLM-serving environments. \textsc{Vortex} combines a Python-embedded frontend language (\vflow), an interpreter that lowers programs into executable \vtensor operators, and an efficient backend integrated with existing serving systems to realize end-to-end sparse attention speedups. We further demonstrated that \textsc{Vortex} enables AI-agent-driven exploration of the sparse attention design space, producing algorithms with strong accuracy--throughput trade-offs.

\section*{Acknowledgements}
We gratefully acknowledge access to NVIDIA computing resources. This work was partially supported by Google Research Award, Google ML \& System Junior Faculty Award, Amazon Research Award, Fireworks AI, Intel, Li Auto, Moffett AI, and CMU CyLab Seed funding. This material is also based upon work supported by the National Science Foundation under Grant Nos. CCF-2504353 and CCF-2247014, and by IARPA. Any opinions, findings, conclusions or recommendations expressed are those of the authors and do not necessarily reflect the views of the National Science Foundation.

\section*{Limitations and Future Work}

\textbf{Limitations.}
\textit{Prefill support.}
Currently, \sys focuses on the decoding stage of LLM serving and does not yet support sparse attention optimization during the prefill phase. While decoding dominates the cost of long-generation workloads, prefill efficiency remains important for many real-world applications, and extending \sys to execute efficient sparse attention during prefill is an important direction.
\textit{Training support.}
\sys is designed for inference serving and targets the forward decoding path. It does not yet support sparse attention during training, where the backward pass introduces additional gradient and memory layout requirements that the current \vtensor abstraction and backend do not address. Supporting sparse attention training would let researchers study algorithms that are co-designed with the training objective rather than applied only at inference time.

\textbf{Future work.}
\textit{RNN and Mamba architectures.}
\sys currently centers on attention-based models with an explicit key-value cache. Extending the \vtensor abstraction to recurrent and state-space architectures such as RNNs and Mamba~\citep{gu2023mamba}, whose state is a fixed-size recurrent representation rather than a growing cache, would broaden \sys to the emerging class of hybrid and linear-attention models~\citep{Zhang2025KimiLA}.
\textit{End-to-end sparse attention algorithm discovery.}
Although \sys enables rapid programming, deployment, and large-scale validation of sparse attention algorithms, it is not itself a fully end-to-end research system. Building end-to-end autonomous systems that jointly propose, analyze, optimize, implementing kernels and validate sparse attention algorithms remains an open research direction.

More broadly, we view \sys as a foundation for future systems that tightly integrate programmable serving infrastructures with AI-driven systems research.

\section*{Broader Impact}
We focus on systems for sparse attention serving in LLMs. While there are numerous application scenarios for large language models that warrant further study of potential societal impact, we would like to highlight that our work does not advance the capabilities of these models.
Our work is primarily an algorithmic study with no specific usage limitations, and while LLMs themselves can be used with malicious purposes, we believe that none of such use cases are specific to this paper.

\section*{Declaration of LLM Usage}

We used large language models (LLMs) to assist with polishing the writing of this paper and to generate illustrative figures unrelated to the experimental results. In addition, this work includes experiments with LLM agents on sparse attention research and optimization.



\clearpage

\bibliographystyle{assets/plainnat}
\bibliography{paper}

\clearpage

\beginappendix
\section{Tensor Layout in LLM Serving}
\label{sec:detailed memory layouts}

Consider a batch of size $b$ with sequence lengths ${s_0,\dots,s_{b-1}}$ and per-token feature shape $\mathbf{h}$.

\textbf{Batch Layout.} A set of independent tensors ${\boldsymbol{x}_i \in \mathbb{R}^{s \times \mathbf{h}} \mid 0 \le i < b}$.

\textbf{Ragged Layout.} A contiguous buffer $\boldsymbol{x}_{\mathrm{flat}} \in \mathbb{R}^{(\sum_i s_i)\times \mathbf{h}}$ with pointer array $\mathbf{p}\in\mathbb{N}^{b+1}$, where $\mathbf{p}[0]=0$ and $\mathbf{p}[i+1]=\mathbf{p}[i]+s_i$. Each sequence is recovered as $\boldsymbol{x}_i=\boldsymbol{x}_{\mathrm{flat}}[\mathbf{p}[i]:\mathbf{p}[i+1]]$.

\textbf{Paged Layout.} With a shared tensor storage $\boldsymbol{x}_{\mathrm{storage}} \in \mathbb{R}^{N \times \mathbf{h}}$, each sequence is represented by a list of page indices $I_i$ and reconstructed as
$\boldsymbol{x}_i=\mathrm{concat}(\boldsymbol{x}_{\mathrm{storage}}[k] \mid k \in I_i)$.
Different sequences may share pages, i.e., $I_i$ and $I_j$ may contain the same indices. The index structure $\mathbf{I}=\{I_0,\dots,I_{b-1}\}$ follows a ragged layout. Both batched and ragged layouts can be constructed as instances of the paged layout.
    

\section{Details of \vtensor}
\label{sec:discussion}

\textit{Shape Propagation.}
For each output tensor $\boldsymbol{y}^{v}_m$, we require all sequences to share the same non-leading dimensions, while allowing the leading dimension to vary across sequences. Formally,
\(
\mathrm{shape}(\boldsymbol{y}^{v}_{m,i}) = (s_{m,i}, \mathbf{h}_m),
\)
where $\mathbf{h}_m$ is identical for all $0 \le i < b$. Under this constraint, the output shape is represented using a ragged layout with pointer array $\mathbf{p}$ satisfying $\mathbf{p}[0]=0$ and $\mathbf{p}[i+1]=\mathbf{p}[i]+s_{m,i}$. The aggregated tensor therefore has shape $\left(\sum_i s_{m,i}\right)\times \mathbf{h}_m$.

\textit{Layout Propagation.}
In addition to shape propagation, each output tensor must also determine its page layout. Specifically, for every sequence $i$, the output tensor $\boldsymbol{y}^{v}_{m,i}$ is associated with a page index list $I_{m,i}$ describing where its data resides in shared storage. These page indices may either be newly allocated or reused from existing tensors, depending on the operator semantics. Therefore, layout propagation is controlled by compiler or user-provided policies that specify how pages are allocated, reused, and propagated across operators. 

\textit{Discussion.} Our abstraction imposes a simple constraint: operators are not allowed to transform the batch dimension. Fortunately, this restriction aligns naturally with practical LLM serving workloads, where sequences within a batch are typically processed independently and cross-sequence operations are rare. By extending PyTorch tensors with layout semantics, \vtensor systematically inherits the semantics of existing PyTorch operators while augmenting them with shape and layout propagation rules. This design enables a simple and compositional programming model for paged tensor computation (see~\Cref{sec:programming model}).

\section{Kernel Templates}
\begin{minipage}{0.96\linewidth}
\begin{lstlisting}[style=custom,caption={Sequence-local operator template},label={lst:seq-local-template}]
function op(input_tensors, output_tensors, winfo_*, indptr, indices):
  pid = program_id(); num_progs = num_programs()
  n = load(winfo_num_workloads)

  per = n // num_progs; r = n % num_progs
  start = pid * per + min(pid, r)
  end = start + per + (pid < r)

  for w in range(start, end):
    b   = load(winfo_batch_indices + w)
    off = load(winfo_offsets + w)
    len = load(winfo_lens + w)

    for q in input_tensors:
      x_q = load_tile(q, b, off, len, indices)

    y_1, ..., y_m = compute(x_1, ..., x_n)

    for q in output_tensors:
      store_tensor(q, y_q, b, off, len, indices)
\end{lstlisting}
\end{minipage}

\section{Ablation Study of \sys Efficiency}
\label{sec:ablations}

In this section, we study two aspects of \sys. First, we analyze the execution breakdown of \sys during a decoding step and quantify the impact of sparse attention kernels. Second, we evaluate the effectiveness of our stochastic radix top-$k$ optimizations.

\subsection{Kernel Profiling.}
\label{sec:profiling}

\begin{figure*}[t]
  \centering
  \includegraphics[width=\linewidth]{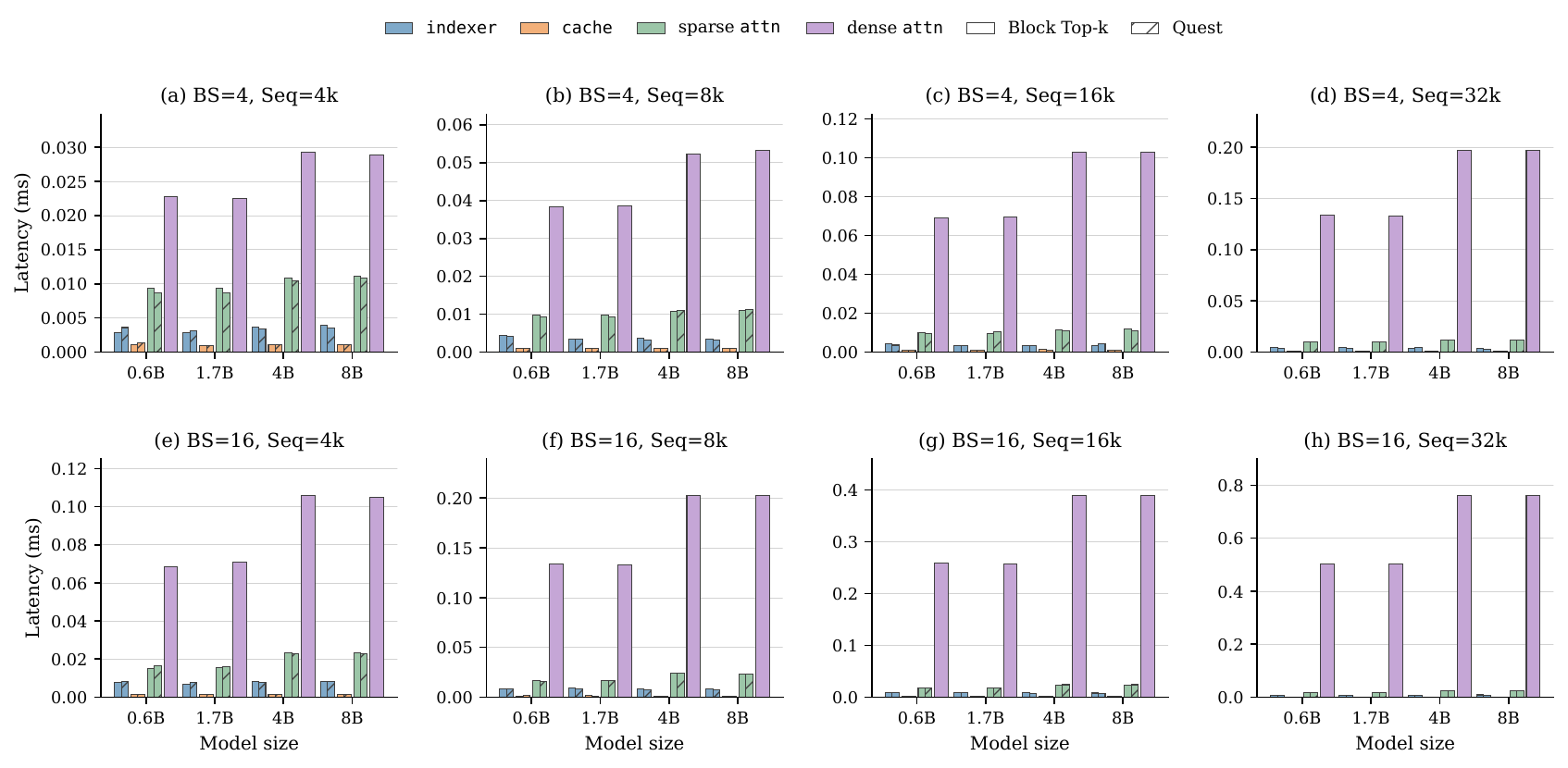}
  \caption{\textbf{Per-kernel decode latency} for dense attention versus two sparse-attention algorithms (block top-$k$~\citep{guo2024blocksparse}, Quest~\citep{tang2024quest}) across batch size, input length, and different sizes of the Qwen3 model. We show the latency of indexer, cache, sparse attention, and dense attention for the same sequence. Note that the indexer and cache kernels are each only $\sim$$1\text{--}10\,\mu$s and are barely visible in the long-context panels.}
  \label{fig:profiling-statistics}
\end{figure*}

Replacing dense attention with sparse attention significantly compresses the decode block latency (\Cref{fig:profiling-statistics}). At BS=16 and 32k context length, block top-$k$ achieves $3.78\text{--}4.81\times$ end-to-end speedup across Qwen3 model sizes, while Quest achieves $3.46\text{--}4.33\times$. At shorter contexts, sparse attention still provides substantial gains, including $2.08\text{--}2.96\times$ at 16k and $1.27\text{--}1.74\times$ at 8k. Even at smaller batch sizes (BS=4, 32k), both methods maintain $1.28\text{--}1.81\times$ speedup.

At the kernel level, sparse attention is substantially faster than dense attention. For example, on Qwen3-8B with BS=16 and 32k context length, the sparse attention kernel requires only $0.025$\,ms compared to $0.760$\,ms for dense attention, yielding over $30\times$ kernel-level acceleration. Meanwhile, indexer and cache-management kernels contribute only a few microseconds, demonstrating that the overhead introduced by \sys is negligible compared to attention computation.

\subsection{Radix Top-$k$ Kernels.}

\textit{Additional Optimization: Remapping.} Before radix partitioning, we apply a monotonic transformation $x'=f(x)$ that preserves ordering while reshaping the value distribution. The transformation improves bin separation and reduces refinement cost. The function $f$ is selected offline through profiling and fused into the histogram pass without additional memory traffic.

\label{sec:topk-kernels}

\begin{figure*}[t]
  \centering
  \includegraphics[width=\linewidth]{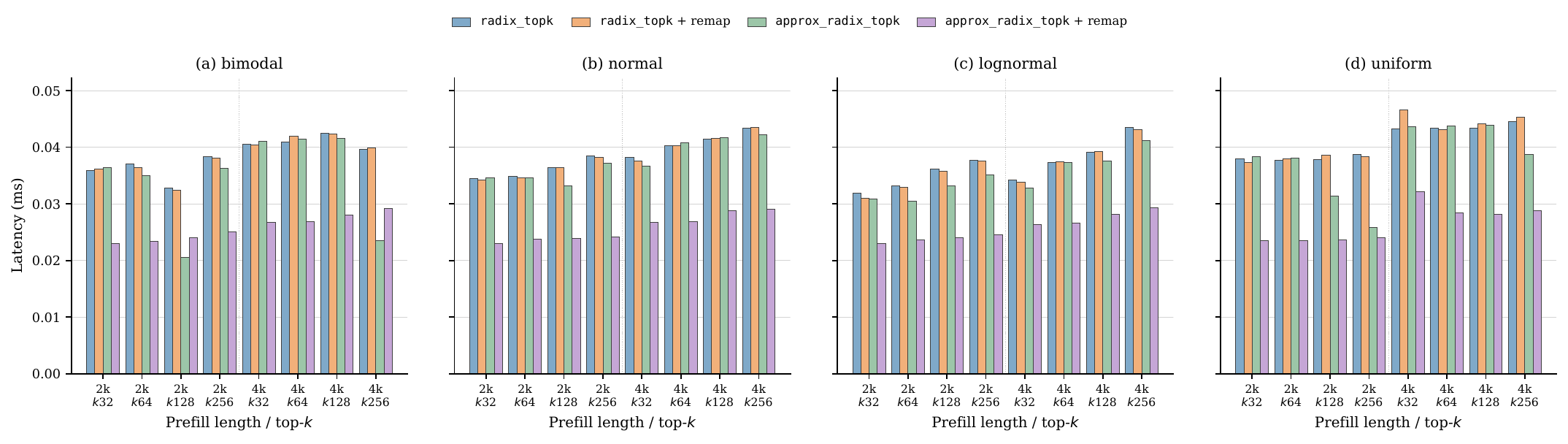}
  \caption{\textbf{Top-$k$ kernel latency} at recall@$k > 0.97$. Mean latency (ms) of our four top-$k$ implementations across (prefill length, top-$k$) pairs, one panel per score distribution: (a) bimodal, (b) normal, (c) lognormal, (d) uniform. We tested on 4 kernels: radix topk, radix topk with remap, radix topk with early termination, and two optimizations combined (\texttt{approx\_radix\_topk + remap}). We report the best variant over remapping and tolerate ratios subject to the recall constraint.}
  \label{fig:topk-kernels}
\end{figure*}

At recall@$k > 0.97$, our \texttt{approx\_radix\_topk + remap} kernel consistently outperforms the \texttt{radix\_topk} baseline across all tested settings (\Cref{fig:topk-kernels}). The speedup ranges from $1.30\text{--}1.62\times$, with an average improvement of $1.49\times$. The gains remain stable across different $k$ values, prefill lengths, and score distributions, including bimodal, normal, lognormal, and uniform distributions. Across all evaluated configurations, \texttt{approx\_radix\_topk + remap} is the fastest implementation among the four compared kernels.

\subsection{Topk Kernels Pareto Analysis}

\begin{figure}[t]
  \centering
  \includegraphics[width=\linewidth]{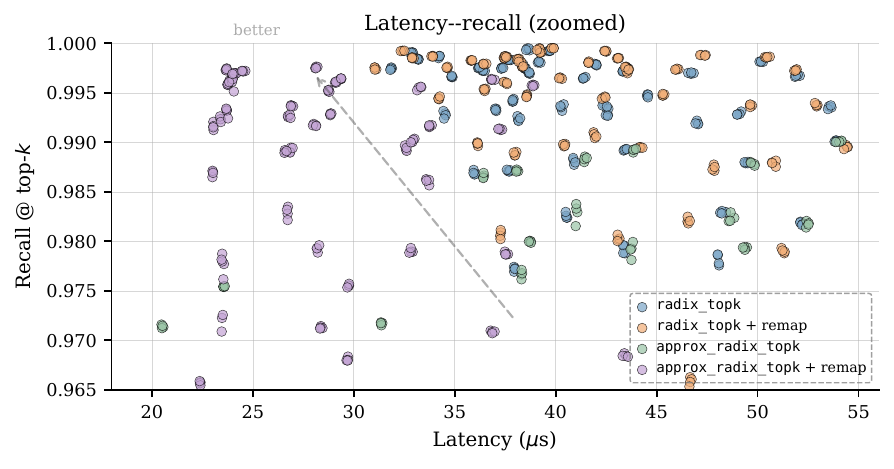}
  \caption{\textbf{Latency--recall Pareto.} Each point is the per-config best-of-family variant (across model sizes, distributions, prefill lengths, and $k$). }
  \label{fig:topk-pareto}
\end{figure}

\paragraph{Pareto coverage.}
The bar comparison in Figure~\ref{fig:topk-kernels} fixes a single recall floor; Figure~\ref{fig:topk-pareto} reports the full latency--recall trade-off so that the choice of operating point is auditable. Each point represents the fastest variant of a kernel family at one (model, distribution, prefill, $k$) cell. The structure of the cloud confirms the reading from Figure~\ref{fig:topk-kernels}: \texttt{approx\_radix\_topk + remap} sits to the \emph{left} of every other family at every recall level above $0.97$, never crossing the \texttt{radix\_topk} or \texttt{approx\_radix\_topk} clouds, so its dominance is not an artifact of the chosen recall floor.

\paragraph{Choice of the recall threshold.}
The operating point is bounded from both sides. \textit{(i) Lower bound from competing variants.}
At thresholds below $\sim$$0.95$, plain \texttt{approx\_radix\_topk} with aggressive tolerate ratios ($\tau \geq 0.25$) becomes eligible and undercuts \texttt{approx\_radix\_topk + remap} in some cells (most visibly on lognormal at small prefill, large $k$), blurring the headline claim. The smallest threshold at which \texttt{approx\_radix\_topk + remap} is uncontested across (prefill, $k$, distribution) is $\sim$$0.96$.
\textit{(ii) Upper bound from kernel feasibility.}
At thresholds above $\sim$$0.99$, even \texttt{approx\_radix\_topk + remap} struggles to qualify in cells where the score histogram is intrinsically diffuse (notably small-$k$ uniform cells, where the maximum reachable recall is $\sim$$0.95$). Operating closer to $1.0$ would force fallback to \texttt{radix\_topk} and eliminate the speedup.

\section{Templates for AI Agents}
\label{sec:templates}
\begin{minipage}{0.96\linewidth}
\begin{lstlisting}[style=custom,caption={Templates for AI Agents},label={lst:Templates for AI Agents}]
schedule_policy: "max(256, 0.0625 * num_blocks)",
skip_layers: [0, 1],
block_size: 16,
page_size: 32,
kv_cache_dtype: "fp8_e4m3",
sparse_attention_file: "my_sparse_attention.py",
sparse_attention_module: "my_sparse_attention"
\end{lstlisting}
\end{minipage}

In \texttt{my\_sparse\_attention.py}, AI agents will register \texttt{my\_sparse\_attention} by sub-classing \vflow. \sys now supports \texttt{bfloat16}, \texttt{fp8\_e4m3} and \texttt{fp8\_e5m2}.

\section{Operators and the Cache/Indexer Flow}
\label{sec:ops}
A \vflow program has two stages. In the \emph{cache stage}, \texttt{create\_cache} declares named auxiliary fields and \texttt{forward\_cache} fills them from the KV cache; this is query-independent and runs per block, so cache-side operators act within a single block. In the \emph{indexer stage}, \texttt{forward\_indexer} consumes the current query together with the cached fields to produce a per-block routing score and ends in a selection op (\texttt{topK}); indexer-side operators act across pages and additionally provide cross-page reductions, normalization, cross-step persistence (\texttt{Load}/\texttt{Save}), and selection. \Cref{tab:ops} lists the operators and the stage(s) in which each is available.

\begin{table}[t]
\centering
\footnotesize
\begin{tabular}{llc}
\toprule
\textbf{Operator} & \textbf{Mathematical meaning} & \textbf{Stage} \\
\midrule
\texttt{Mean}, \texttt{Max}, \texttt{Min}, \texttt{L2Norm} & average / max / min / $\sqrt{\sum_i x_i^2}$ along an axis & Both \\
\texttt{Sum} & sum along an axis (including cross-page, \texttt{dim}$=0$) & Indexer \\
\texttt{*Interleave} & the same reductions over each group of $k$ consecutive entries & Cache \\
\addlinespace
\texttt{GeMM} & block- or page-wise matrix product $Y X^{\top}$ & Both \\
\texttt{Multiply}, \texttt{Add} & $X \odot Y$ and $\alpha X + \beta Y$ (broadcast) & Both \\
\texttt{Maximum}, \texttt{Minimum} & $\max(X,Y)$ and $\min(X,Y)$ & Both \\
\texttt{Kron} & Kronecker product over inner axes & Indexer \\
\addlinespace
\texttt{Relu}, \texttt{Sigmoid}, \texttt{Silu} & standard activations of $\beta x+\alpha$ & Both \\
\texttt{Exp}, \texttt{Log}, \texttt{Abs}, \texttt{Add\_Mul} & $e^{\beta x+\alpha}$, $\log(\beta x+\alpha)$, $|\beta x+\alpha|$, $\beta x+\alpha$ & Both \\
\texttt{Where\{Eq,Ne,Gt,Ge,Lt,Le\}} & $0$ where the comparison holds and $-\infty$ otherwise & Both \\
\addlinespace
\texttt{Softmax}, \texttt{Normalize} & $\mathrm{softmax}(s\,X)$ and $X/\sum X$ along an axis & Indexer \\
\texttt{Conv1d} & 1-D convolution along an axis with a given kernel & Indexer \\
\addlinespace
\texttt{Reshape} & same-size reshape & Both \\
\texttt{Transpose} & swap the last two axes & Indexer \\
\texttt{MaskSlice} & $\alpha$ inside a position window $[s,e)$ and $\beta$ outside & Both \\
\texttt{Fill}, \texttt{CFill} & write a constant to all entries (\texttt{CFill} zeroes a saved field) & Cache \\
\addlinespace
\texttt{Load}, \texttt{Save} & read / write a named field across decoding steps & Indexer \\
\texttt{topK} & exact top-$k$ blocks by score & Indexer \\
\texttt{approxTopK} & adaptive radix top-$k$ ($\texttt{tolerate\_ratio}\in[0,1]$ trades exactness for speed) & Indexer \\
\texttt{TopK}, \texttt{Union} & block-table selection and its deduplicated union (trtllm backend) & Indexer \\
\bottomrule
\end{tabular}
\caption{Representative \vflow operators and their mathematical meaning. \emph{Stage} indicates availability on the cache side (within a block) and/or the indexer side (across pages).}
\label{tab:ops}
\end{table}
\section{AI-Agent-Proposed Algorithms}
\label{sec:agent-algos}
Each model proposed $20$ sparse-attention flows in the one-shot setting of \Cref{sec:innovation}. We describe all of them below in words and notation rather than code.

\paragraph{The shared pipeline.} Every flow is a complete instance of the same four-stage computation that, at each decoding step, turns the query $q$ and the KV cache $(K,V)$ into an attention output. \textbf{(1) Build cache.} The KV sequence is split into blocks of size $b$; block $j$ holds keys $K_j$ and values $V_j$ with token vectors $k_t,v_t$, from which the flow precomputes (in \texttt{create\_cache}/\texttt{forward\_cache}) the per-block statistics it will need, such as the key centroid $c_j=\tfrac1b\sum_{t\in j}k_t$, the value centroid $\bar v_j$, the feature-wise key envelope $k^{\max}_{j},k^{\min}_{j}$, the mean token value norm $\overline{\lVert v\rVert}_j$, the mean key norm $\overline{\lVert k\rVert}_j$, or a set of finer sub-centroids $c_j^{(g)}$. \textbf{(2) Form the query view.} The decode query is reduced to a routing query, either the group-averaged query $\bar q$ or a per-head query $q_h$. \textbf{(3) Score.} The flow computes a scalar score $s_j$ for every block from the query view and the cached statistics; this score is the only stage that differs across methods. \textbf{(4) Select and attend.} The top-$k$ blocks by $s_j$ are kept and the query runs \emph{exact} softmax attention over only those blocks' keys and values, producing the output. Flows that name an exponential moving average (EMA) or a previous-step quantity carry state across decoding steps through \texttt{Load}/\texttt{Save}; a few customize stage~(4) (for example masking with $-\infty$ instead of a fixed top-$k$), which we note inline.

For each algorithm we therefore give only what distinguishes it, namely the block statistics it caches and the score it computes; stages~(2) and~(4) are as above unless stated. Most ideas recombine a few recurring motifs, namely centroid alignment, the Quest key envelope, value-energy gating, multi-resolution sub-blocks, and temporal accumulation.

\paragraph{Claude Opus 4.7.}
\algname{Score velocity} \textit{Cache:} key centroid $c_j$ and the previous step's score $s_j^{\mathrm{prev}}$ (carried via \texttt{Save}/\texttt{Load}). \textit{Score:} $s_j=\langle\bar q,c_j\rangle-\tfrac12 s_j^{\mathrm{prev}}$. Rewards blocks whose relevance is rising and discounts ones already heavily attended.

\algname{Head-consensus Quest} \textit{Cache:} key envelope $k^{\max}_j,k^{\min}_j$. \textit{Score:} $s_j=\min_h\sum_d\max(q_h k^{\max}_{j,d},q_h k^{\min}_{j,d})$, evaluating the Quest envelope per head and keeping a block only if it looks relevant to every head.

\algname{Multi-resolution agreement} \textit{Cache:} several sub-centroids $c_j^{(g)}$ per block. \textit{Score:} $s_j=\min_g\langle\bar q,c_j^{(g)}\rangle$, the least-aligned sub-region, so a block is kept only if it matches the query at every resolution.

\algname{Pessimistic envelope} \textit{Cache:} key envelope $k^{\max}_j,k^{\min}_j$. \textit{Score:} $s_j=\max_h\sum_d\min(q_h k^{\max}_{j,d},q_h k^{\min}_{j,d})$, inverting Quest to a conservative lower bound on similarity.

\algname{Value-energy gate} \textit{Cache:} key centroid $c_j$ and mean value norm $\overline{\lVert v\rVert}_j$. \textit{Score:} $s_j=\langle\bar q,c_j\rangle\cdot\overline{\lVert v\rVert}_j$, favoring blocks whose values carry more energy.

\algname{Volatility boost} \textit{Cache:} centroid $c_j$ and a running mean and second moment of the block score (state). \textit{Score:} $s_j=\langle\bar q,c_j\rangle+\tfrac12\widehat{\mathrm{Var}}_j$, periodically re-examining blocks whose relevance fluctuates.

\algname{Anti-redundancy} \textit{Cache:} centroid $c_j$ and an EMA of past scores (state). \textit{Score:} $s_j=\langle\bar q,c_j\rangle-0.4\,\mathrm{EMA}[s_j]$, discouraging repeated selection of the same blocks.

\algname{Positional smoothing} \textit{Cache:} centroid $c_j$. \textit{Score:} the base alignment $\langle\bar q,c_j\rangle$ convolved along the sequence axis with a $[0.25,0.5,0.25]$ kernel, damping isolated spikes and rewarding locally coherent regions.

\algname{Saturated features} \textit{Cache:} centroid $c_j$. \textit{Score:} $s_j=\sum_d\sigma(\bar q_d c_{j,d})$, passing each feature contribution through a sigmoid so no single dimension dominates.

\algname{Magnitude only} \textit{Cache:} mean key norm $\overline{\lVert k\rVert}_j$. \textit{Score:} $s_j=\overline{\lVert k\rVert}_j$, a query-independent baseline that attends to high-energy regions regardless of the query.

\algname{Value-centroid routing} \textit{Cache:} value centroid $\bar v_j$. \textit{Score:} $s_j=\langle\bar q,\bar v_j\rangle$, testing whether value content alone is a useful routing signal.

\algname{Dual-signal product} \textit{Cache:} key centroid $c_j$ and per-feature key norms $\lVert k_{\cdot,d}\rVert_j$. \textit{Score:} $s_j=\langle\bar q,c_j\rangle\cdot\sum_d\bar q_d\lVert k_{\cdot,d}\rVert_j$, requiring both directional and magnitude agreement.

\algname{Moderate-relevance picker} \textit{Cache:} centroid $c_j$. \textit{Score:} the centroid alignment multiplied by an inverted softmax of itself, suppressing the largest scores so selection peaks at moderate rather than extreme relevance.

\algname{Head $\times$ region consensus} \textit{Cache:} sub-centroids $c_j^{(g)}$. \textit{Score:} $s_j=\min_{h,g}\langle q_h,c_j^{(g)}\rangle$ over all head-by-sub-centroid pairings (a Kronecker grid), demanding agreement across both heads and intra-block regions.

\algname{Envelope spread} \textit{Cache:} key envelope $k^{\max}_j,k^{\min}_j$. \textit{Score:} $s_j=\sum_d q_d(k^{\max}_{j,d}-k^{\min}_{j,d})$, preferring blocks with a wide per-feature key range.

\algname{Smoothed query} \textit{Cache:} centroid $c_j$ and the previous routing query $\bar q^{\mathrm{prev}}$ (state). \textit{Score:} $s_j=\langle 0.7\,\bar q^{\mathrm{prev}}+0.3\,\bar q,\,c_j\rangle$, stabilizing selection against per-step query noise.

\algname{Elevation over history} \textit{Cache:} centroid $c_j$ and a running minimum of the block score (state). \textit{Score:} $s_j=\langle\bar q,c_j\rangle-\min\text{-hist}_j$, selecting blocks currently well above their own historical floor.

\algname{Coarse $\times$ fine} \textit{Cache:} whole-block centroid $c_j$ and half-block sub-centroids $c_j^{(g)}$. \textit{Score:} $s_j=\langle\bar q,c_j\rangle\cdot\max_g\langle\bar q,c_j^{(g)}\rangle$, combining a coarse and a fine view of the block.

\algname{Key-and-value agreement} \textit{Cache:} key centroid $c_j$ and value centroid $\bar v_j$. \textit{Score:} $s_j=\langle\bar q,c_j\rangle\cdot\langle\bar q,\bar v_j\rangle$, keeping blocks where both representations point toward the query.

\algname{Peak sharpening} \textit{Cache:} centroid $c_j$. \textit{Score:} $\langle\bar q,c_j\rangle$ convolved with a $[-0.5,1.5,-0.5]$ kernel, isolating sharp peaks and penalizing blocks surrounded by similarly high neighbors.

\paragraph{Claude Sonnet 4.6.}
\algname{Value-energy gate} \textit{Cache:} key centroid $c_j$ and mean value norm $\overline{\lVert v\rVert}_j$. \textit{Score:} $s_j=\langle\bar q,c_j\rangle\cdot\overline{\lVert v\rVert}_j$, favoring high-energy blocks.

\algname{Key-value consensus} \textit{Cache:} key centroid $c_j$ and value centroid $\bar v_j$. \textit{Score:} $s_j=\min(\langle\bar q,c_j\rangle,\langle\bar q,\bar v_j\rangle)$, so a block must agree on both to score highly.

\algname{Value-weighted centroid} \textit{Cache:} a value-norm-weighted key centroid $c_j^{w}=\big(\sum_{t\in j}\lVert v_t\rVert\,k_t\big)/\big(\sum_{t\in j}\lVert v_t\rVert\big)$ instead of a plain mean. \textit{Score:} $s_j=\langle\bar q,c_j^{w}\rangle$, so keys paired with high-energy values dominate the block representation.

\algname{Distance routing} \textit{Cache:} key centroid $c_j$. \textit{Score:} $s_j=-\lVert\bar q-c_j\rVert_1$, selecting blocks closest under an absolute-difference metric.

\algname{Magnitude co-activation} \textit{Cache:} mean absolute key $\mathbb{E}[|k|]_j$. \textit{Score:} $s_j=\langle|\bar q|,\mathbb{E}[|k|]_j\rangle$, matching on activation strength rather than sign.

\algname{Rectified Quest} \textit{Cache:} key envelope $k^{\max}_j,k^{\min}_j$. \textit{Score:} $s_j=\max_h\sum_d\mathrm{ReLU}(q_h k^{\max}_{j,d})$, summing only positive envelope evidence.

\algname{Head $\times$ region peaks} \textit{Cache:} sub-centroids $c_j^{(g)}$. \textit{Score:} $s_j=\max_{h,g}\langle q_h,c_j^{(g)}\rangle$, so any strongly matching head-region pair can select a block.

\algname{Smoothly gated features} \textit{Cache:} centroid $c_j$. \textit{Score:} $s_j=\sum_d\mathrm{SiLU}(\bar q_d c_{j,d})$, a soft alternative to hard feature thresholding.

\algname{Magnitude bias} \textit{Cache:} centroid $c_j$ and its norm $\lVert c_j\rVert$. \textit{Score:} $s_j=\langle\bar q,c_j\rangle+0.1\log\lVert c_j\rVert$, gently preferring blocks with larger key magnitude.

\algname{Above-mean gate} \textit{Cache:} centroid $c_j$. \textit{Score:} $\langle\bar q,c_j\rangle$, but \textit{select} adaptively, keeping a block only when its score exceeds the per-step mean across blocks and masking the rest with $-\infty$ rather than taking a fixed top-$k$.

\algname{Positional smoothing} \textit{Cache:} centroid $c_j$. \textit{Score:} $\langle\bar q,c_j\rangle$ smoothed by a 3-tap kernel along the sequence axis, rewarding locally coherent regions.

\algname{Persistent value energy} \textit{Cache:} centroid $c_j$ and an EMA (rate $0.95$) of the block value norm (state). \textit{Score:} $s_j=\langle\bar q,c_j\rangle\cdot\mathrm{EMA}[\overline{\lVert v\rVert}_j]$, a slowly varying value-importance weight.

\algname{Two-timescale agreement} \textit{Cache:} centroid $c_j$ and a fast and a slow EMA of the score (state). \textit{Score:} the minimum of the two EMAs, keeping blocks relevant on both short and long horizons.

\algname{Query-weighted spread} \textit{Cache:} per-sub-block key envelopes $k^{\max}_{j,g},k^{\min}_{j,g}$. \textit{Score:} $s_j=\max_g\langle\bar q,\,k^{\max}_{j,g}-k^{\min}_{j,g}\rangle$, combining the Quest envelope with sub-block resolution.

\algname{Entropy contribution} \textit{Cache:} centroid $c_j$. \textit{Score:} $s_j=p_j(-\log p_j)$ with $p=\mathrm{softmax}_j\langle\bar q,c_j\rangle$, emphasizing mid-probability blocks over both certain and negligible ones.

\algname{Curvature bonus} \textit{Cache:} centroid $c_j$. \textit{Score:} $s_j=\langle\bar q,c_j\rangle+0.5\,\big|[-1,2,-1]*\langle\bar q,c\rangle\big|_j$, boosting blocks at local extrema of the score profile.

\algname{Value-only routing} \textit{Cache:} value centroid $\bar v_j$. \textit{Score:} $s_j=\langle\bar q,\bar v_j\rangle$, a check on how much routing signal lives in the values.

\algname{Gated key-value product} \textit{Cache:} key centroid $c_j$ and value centroid $\bar v_j$. \textit{Score:} $s_j=\mathrm{ReLU}(\langle\bar q,c_j\rangle)\cdot\mathrm{ReLU}(\langle\bar q,\bar v_j\rangle)$, a smooth logical AND of the two signals.

\algname{Conservative envelope} \textit{Cache:} lower key envelope $k^{\min}_j$. \textit{Score:} $s_j=\max_h\sum_d q_h k^{\min}_{j,d}$, a pessimistic counterpart to standard Quest.

\algname{Sigmoid-gated Quest} \textit{Cache:} key envelope $k^{\max}_j,k^{\min}_j$. \textit{Score:} $s_j=\max_h\sum_d\sigma\!\big(q_d(k^{\max}_{j,d}-k^{\min}_{j,d})\big)\max(q_d k^{\max}_{j,d},q_d k^{\min}_{j,d})$, down-weighting features whose envelope is narrow and thus uninformative.

\paragraph{GPT-5.}
\algname{Feature value-temperature} \textit{Cache:} key centroid $c_j$ and per-feature value norms $\lVert v_{\cdot,d}\rVert_j$. \textit{Score:} $s_j=\sum_d(\bar q_d c_{j,d})\,\lVert v_{\cdot,d}\rVert_j$, letting value energy act as a per-feature temperature on key relevance.

\algname{Value-scaled Quest} \textit{Cache:} key envelope $k^{\max}_j,k^{\min}_j$ and per-feature value norms. \textit{Score:} the head-maximum of the Quest envelope products with each feature scaled by $\lVert v_{\cdot,d}\rVert_j$, combining envelope key matching with value importance.

\algname{Magnitude-penalized alignment} \textit{Cache:} centroid $c_j$ and its absolute value $|c_j|$. \textit{Score:} $s_j=\langle\bar q,c_j\rangle-0.25\,\langle\bar q,|c_j|\rangle$, penalizing blocks that score highly only through large-magnitude features.

\algname{Sequence value gate} \textit{Cache:} key centroid $c_j$ and value centroid $\bar v_j$. \textit{Score:} $s_j=\langle\bar q,c_j\rangle+\mathbf{1}\!\left[\langle\bar q,\bar v_j\rangle>\overline{\langle\bar q,\bar v\rangle}\right]$, a binary value-side gate atop centroid routing.

\algname{Persistent value-weighted score} \textit{Cache:} $c_j$, per-feature value norms, and an EMA accumulator $r_j$ (state). \textit{Score:} $r_j\leftarrow0.7\,r_j+\sum_d(\bar q_d c_{j,d})\lVert v_{\cdot,d}\rVert_j$ with $s_j=r_j$, smoothing the value-weighted signal over decoding steps.

\algname{Sub-block peak} \textit{Cache:} four sub-centroids $c_j^{(g)}$. \textit{Score:} $s_j=\max_g\langle\bar q,c_j^{(g)}\rangle$, so a strong match in any region selects the block.

\algname{Sub-block average} \textit{Cache:} four sub-centroids $c_j^{(g)}$. \textit{Score:} $s_j=\tfrac14\sum_g\langle\bar q,c_j^{(g)}\rangle$, a smoother counterpart to the peak version.

\algname{Interleaved envelope} \textit{Cache:} the interleaved key envelope. \textit{Score:} the maximum over a Kronecker pairing of the query with the interleaved envelope, exposing intra-block envelope structure to the score.

\algname{Outlier value gate} \textit{Cache:} centroid $c_j$ and mean value norm $\overline{\lVert v\rVert}_j$. \textit{Score:} $s_j=\langle\bar q,c_j\rangle+\mathbf{1}[\,\bar q\!\cdot\!\overline{\lVert v\rVert}_j>\text{mean}\,]$, promoting value outliers.

\algname{Softmax value temperature} \textit{Cache:} centroid $c_j$ and mean value norm $\overline{\lVert v\rVert}_j$. \textit{Score:} $s_j=\mathrm{softmax}_j(\langle\bar q,c_j\rangle)\cdot(\bar q\!\cdot\!\overline{\lVert v\rVert}_j)$, sharpening relevant blocks while weighting by value content.

\algname{Sigmoid value gate} \textit{Cache:} centroid $c_j$ and mean value norm $\overline{\lVert v\rVert}_j$. \textit{Score:} $s_j=\langle\bar q,c_j\rangle\cdot\sigma(-0.02\,\bar q\!\cdot\!\overline{\lVert v\rVert}_j)$, a soft saturating value modulation.

\algname{Lower-envelope penalty} \textit{Cache:} centroid $c_j$ and lower key envelope $k^{\min}_j$. \textit{Score:} $s_j=\langle\bar q,c_j\rangle-0.5\,\langle\bar q,k^{\min}_j\rangle$, penalizing blocks whose minimum keys already align with the query.

\algname{Key/value blend} \textit{Cache:} key centroid $c_j$ and value centroid $\bar v_j$. \textit{Score:} $s_j=0.75\,\langle\bar q,c_j\rangle+0.25\,\langle\bar q,\bar v_j\rangle$, a soft combination of the two signals.

\algname{Key-value product} \textit{Cache:} key centroid $c_j$ and value centroid $\bar v_j$. \textit{Score:} $s_j=\langle\bar q,c_j\rangle\cdot\langle\bar q,\bar v_j\rangle$, requiring both to be high at once.

\algname{Persistent surprise} \textit{Cache:} centroid $c_j$ and an EMA accumulator $r_j$ (state). \textit{Score:} $r_j\leftarrow0.6\,r_j+\big|\langle\bar q,c_j\rangle-\mathrm{mean}_{j'}\langle\bar q,c_{j'}\rangle\big|$ with $s_j=r_j$, selecting blocks that are persistently atypical.

\algname{Head-peak value scale} \textit{Cache:} per-head key centroid $c_j$ and mean value norm $\overline{\lVert v\rVert}_j$. \textit{Score:} $s_j=\max_h\langle q_h,c_j\rangle\cdot(\bar q\!\cdot\!\overline{\lVert v\rVert}_j)$, scaling the strongest head match by block value energy.

\algname{Value-only energy} \textit{Cache:} mean value norm $\overline{\lVert v\rVert}_j$. \textit{Score:} $s_j=\bar q\!\cdot\!\overline{\lVert v\rVert}_j$, a baseline using no key statistics.

\algname{Key-only energy} \textit{Cache:} mean key norm $\overline{\lVert k\rVert}_j$. \textit{Score:} $s_j=\bar q\!\cdot\!\overline{\lVert k\rVert}_j$, the key-side counterpart using no value statistics.

\algname{Positional smoothing} \textit{Cache:} centroid $c_j$. \textit{Score:} $\langle\bar q,c_j\rangle$ with a triangular 3-tap smoothing along the sequence axis.

\algname{Normalized relevance $\times$ value} \textit{Cache:} centroid $c_j$ and mean value norm $\overline{\lVert v\rVert}_j$. \textit{Score:} the product of a normalized centroid-relevance distribution and a normalized value-energy distribution, selecting blocks that rank highly on both.

\end{document}